\documentclass[10pt,twocolumn,letterpaper]{article}

\usepackage[pagenumbers]{cvpr} 

\usepackage[dvipsnames]{xcolor}

\usepackage{pifont}
\usepackage{multirow}
\usepackage{amssymb}
\newcommand*{\ourmodel}{VeloDepth\@\xspace}
\newcommand{\cmark}{\ding{51}}
\newcommand{\xmark}{\ding{55}}
\newcommand{\best}[1]{\mathbf{#1}}
\newcommand{\scnd}[1]{\underline{#1}}
\newcommand{\PAR}[1]{\noindent{\bf #1}}

\definecolor{cvprblue}{rgb}{0.21,0.49,0.74}
\usepackage[pagebackref,breaklinks,colorlinks,citecolor=cvprblue]{hyperref}

\usepackage{microtype}
\def\paperID{61} 
\def\confName{3DV\xspace}
\def\confYear{2026\xspace}

\title{Video Depth Propagation}

\author{
Luigi Piccinelli\textsuperscript{*}\textsuperscript{1} \quad Thiemo Wandel\textsuperscript{*}\textsuperscript{1} \quad Christos Sakaridis\textsuperscript{1} \quad  
Wim Abbeloos\textsuperscript{2} \quad Luc Van Gool\textsuperscript{1,3}\\[0.4cm]
$^1$ETH Z\"urich \quad $^2$Toyota Motor Europe \quad $^3$INSAIT, Sofia University St. Kliment Ohridski
}

\begin{document}

\maketitle

\begingroup
\renewcommand\thefootnote{\fnsymbol{footnote}}
\footnotetext[1]{Denotes equal contribution.}
\endgroup

\begin{abstract}
Depth estimation in videos is essential for visual perception in real-world applications.
However, existing methods either rely on simple frame-by-frame monocular models, leading to temporal inconsistencies and inaccuracies, or use computationally demanding temporal modeling, unsuitable for real-time applications.
These limitations significantly restrict general applicability and performance in practical settings.
To address this, we propose \emph{\ourmodel}, an efficient and robust online video depth estimation pipeline that effectively leverages spatiotemporal priors from previous depth predictions and performs deep feature propagation.
Our method introduces a novel Propagation Module that refines and propagates depth features and predictions using flow-based warping coupled with learned residual corrections.
In addition, our design structurally enforces temporal consistency, resulting in stable depth predictions across consecutive frames with improved efficiency.
Comprehensive zero-shot evaluation on multiple benchmarks demonstrates the state-of-the-art temporal consistency and competitive accuracy of \ourmodel, alongside its significantly faster inference compared to existing video-based depth estimators.
\ourmodel thus provides a practical, efficient, and accurate solution for real-time depth estimation suitable for diverse perception tasks. Code and models are available at \href{https://github.com/lpiccinelli-eth/velodepth}{github.com/lpiccinelli-eth/velodepth}.
\end{abstract}    
\section{Introduction}
\label{sec:introduction}

\begin{figure}[ht]
    \centering
    \footnotesize
    \includegraphics[width=1.0\linewidth]{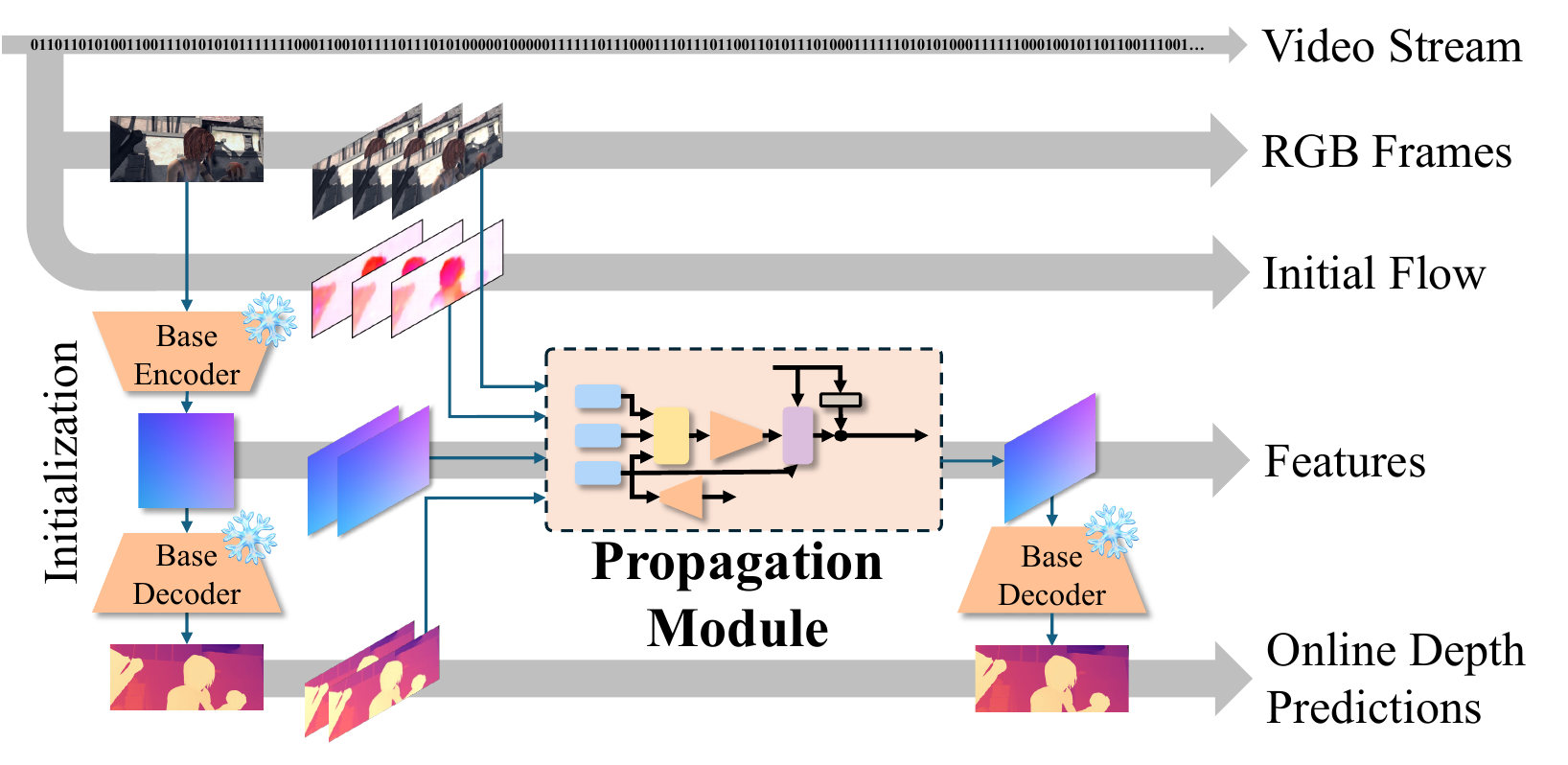}
    \vspace{-10pt}
    \caption{\textbf{\ourmodel} learns to leverage prior information contained in video data, such as previous predictions and scene dynamics. The Propagation Module refines the previous frame features, which ``Base Decoder'' decodes as current frame predictions. The module also propagates the features, along with the predictions, as the next frame's inputs. The prior information and the propagation lead to improved consistency and more efficient inference while maintaining the per-frame performance of the large ``Base Model''.
    }
    \label{fig:method:teaser}
\end{figure}

Depth estimation is a fundamental task in computer vision, which enables a dense perception of the geometric structure of the surrounding scene that is pivotal in a vast variety of applications ranging from autonomous systems~\cite{wang2019depth4vehicles, park2021dd3d} and robotics~\cite{Zhou2019, dong2022depth4robotics} to augmented reality~\cite{deng2022nerf} and medicine~\cite{liu2019medicine}.
While the basic monocular setting of this task, \ie monocular depth estimation (MDE) from single images, is inherently ill-posed due to scale ambiguity and offers fewer priors to learn, its simplicity has historically led to far more attention than depth estimation from videos, especially in the deep learning era~\cite{Eigen2014, Yuan2022newcrf, piccinelli2023idisc, piccinelli2024unidepth, yin2023metric3d, ranftl2020midas, yang2024da1, bochkovskii2024depthpro}.

However, the video setting is better constrained since video sequences inherently provide strong priors, unlike single images, which can be leveraged to improve depth estimation.
In particular, consecutive frames contain redundant visual information, allowing previous depth predictions and features to serve as informative cues for future frames.
Even when estimated approximately, motion provides additional constraints on depth evolution over time.
Leveraging temporal priors by propagating features and depth estimates across frames should lead to more accurate, consistent, and computationally efficient video-based depth methods.
However, existing methods either ignore these priors, \ie MDE, leading to temporal inconsistencies and flickering artifacts, or rely on computationally expensive solutions such as test-time optimization~\cite{Luo2020cvd, kopf2021rcvd}, full temporal attention~\cite{wang2023nvds, chen2025vda}, or video diffusion~\cite{shao2024chrono, hu2024crafter}, making them impractical for real-time applications.
Moreover, these methods usually require future frames not only during training but also during inference, rendering them impractical for most real-world applications, which typically need to run online.

We respond to the above shortcomings in the literature by proposing \ourmodel, a video metric depth estimator that is based on the propagation of depth-related information in a video across frames through time.
The core principle of our approach is to exploit depth predictions and feature representations from previous frames, using them as informative priors to bootstrap the computation for subsequent frames.
Specifically, our method employs a temporal propagation strategy where previously computed depth features and outputs are warped forward through fast but inaccurate optical flow and then refined via a learned residual correction, as depicted in \cref{fig:method:teaser}.
This design structurally enforces temporal consistency, as depth estimation at each frame inherently benefits from previously estimated features and outputs.
Moreover, our approach enhances computational efficiency, since the propagation module only needs to learn the simpler residual mapping from propagated features, rather than performing a full RGB-to-depth prediction from scratch for every frame.
Therefore, \ourmodel~achieves comparable accuracy to computationally intensive single-image models applied frame-by-frame, while simultaneously largely increasing consistency and presenting the efficiency required for real-time applications.

We validate our approach with extensive experiments across four diverse benchmarks, demonstrating its robustness under different motion and scene conditions.
Our results show that leveraging spatiotemporal priors leads to a better trade-off between temporal stability, computational efficiency, and overall depth accuracy compared to standard monocular depth models and prior video-based depth approaches.

\section{Related Works}
\label{sec:relwork}

\PAR{Monocular depth estimation} was proposed in its end-to-end neural network formulation in~\cite{Eigen2014}.
However, monocular methods~\cite{Eigen2014, Laina2016, regressionforest, liu2014deepconvolutionalneuralfields, Fu2018Dorn, Yuan2022newcrf, Patil2022p3depth, piccinelli2023idisc} typically suffer from generalization issues due to limited data and the inherently ill-posed nature of monocular 2D-to-3D unprojection.
Affine-invariant (relative) depth estimation mitigates this by predicting depth up to an unknown scale and shift, removing ill-posedness and improving cross-dataset performance~\cite{ranftl2020midas, yang2024da1, ke2024marigold}.
However, relative depth estimation is unsuitable for physical, metric applications.
More recent works strive for generalizable metric MDE incorporating camera information into the input~\cite{yin2023metric3d, hu2024metric3dv2}, internal features~\cite{piccinelli2024unidepth, piccinelli2025unidepthv2, piccinelli2025unik3d}, or output space~\cite{bochkovskii2024depthpro}.
All MDE methods increase both data and compute to improve performance at the cost of real-time feasibility.
Moreover, they are inherently trained in an image-based fashion, ignoring any temporal information and leading to inconsistencies across frames when run on videos.

\PAR{Offline video depth estimation} leverages all frames of the input video to enhance both temporal performance and accuracy over single-frame depth estimation.
The paradigm defined by~\cite{Luo2020cvd, kopf2021rcvd} involves test-time optimization on initial depth estimates with either fixed or optimizable camera poses.
\cite{shao2024chrono, hu2024crafter} have been the first to repurpose video diffusion models for video depth estimation, while Video Depth Anything~\cite{chen2025vda} extends a large pre-trained affine MDE~\cite{yang2024da2} by incorporating a spatiotemporal head that uses attention to correlate information across frames.
However, these methods suffer from significant drawbacks, including high memory consumption and the inability to produce metric depth predictions.
Moreover, their superior temporal consistency can be attributed to their offline nature, i.e.\ processing videos in chunks, where future frames are also included.
On the other hand, \ourmodel does not require processing the entire video for each frame, which renders our method online, efficient, as well as capable of providing high consistency.

\PAR{Online video depth estimation} aims at online and possibly real-time, temporally consistent depth estimation.
Early methods relied on recurrent architectures, such as LSTM~\cite{sepp1997lstm}, to retain temporal features~\cite{zhang2019cslstm, lstm_mde, cs2018depthnet}, while others incorporated LiDAR for multi-modal fusion~\cite{Patil2020DontFT} or introduced stabilization networks to refine external depth~\cite{wang2023nvds, NVDSPLUS}.
Most of the above methods are based on recurrent networks to retain past features but suffer from drift, vanishing gradients, and a poor capacity-efficiency tradeoff, which limits their effectiveness on real-time and long sequences.
Stabilization-based methods~\cite{wang2023nvds, NVDSPLUS} refine depth estimates post-prediction but introduce additional computational overhead and fail to fully leverage past information.
Yasarla \textit{et al.}~\cite{yasarla2025mamo} proposed an optical flow-based attention memory, which ignores any features from previous predictions or flow, although requiring high-quality flow, and exploits memory-intensive attention.
\ourmodel avoids these pitfalls by directly incorporating the previous frame's ``neck features'', depth predictions, and optical flow as priors, which, combined with a strong initialization, ensures consistency while maintaining computational efficiency.

\section{Method} \label{sec:method}

\begin{figure*}[ht]
    \centering
    \footnotesize
    \includegraphics[width=1.0\linewidth]{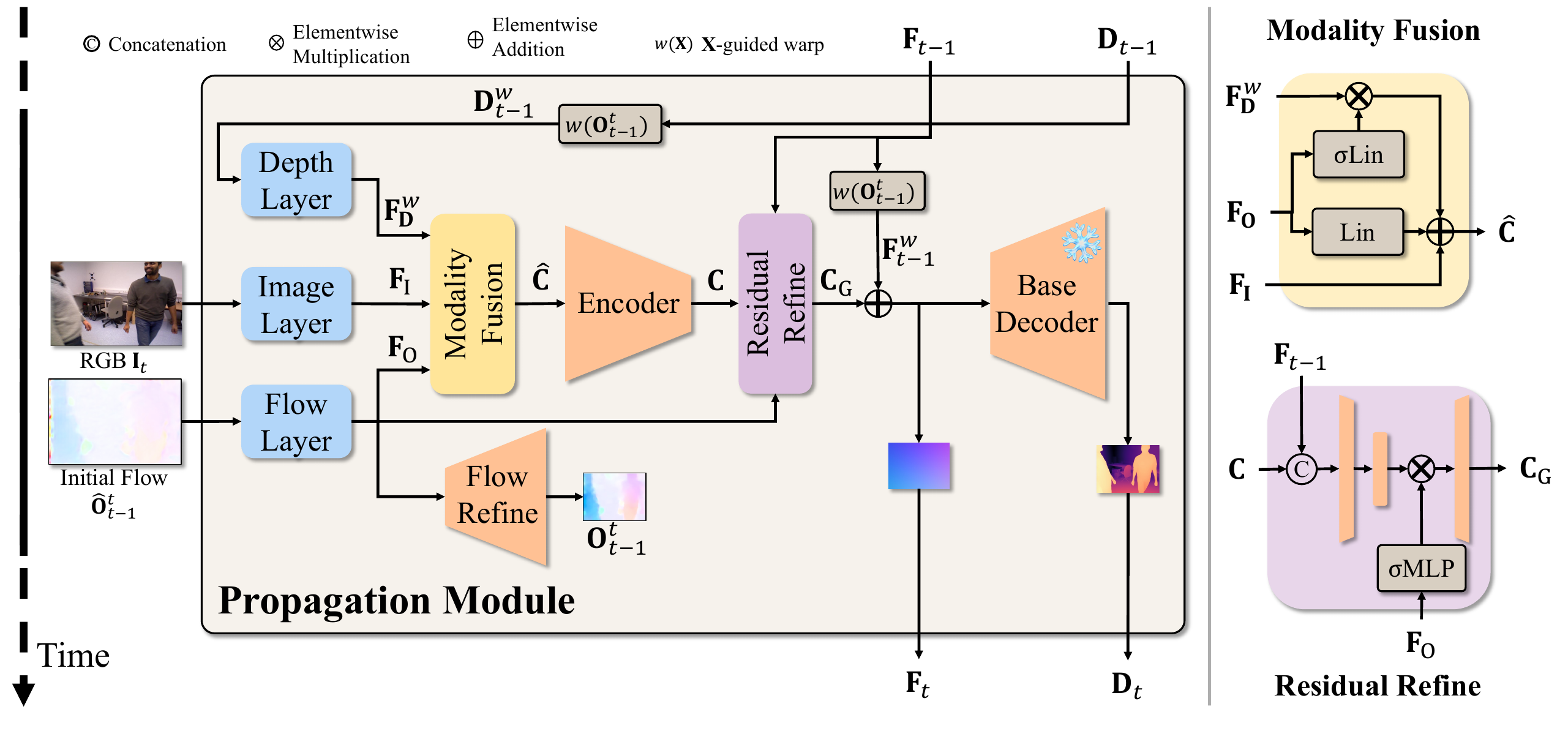}
    \vspace{-14pt}
    \caption{\textbf{Model architecture.} \ourmodel's Propagation Module is fed with the depth prediction from the previous frame ($\mathbf{D}_{t-1}$) and respective features ($\mathbf{F}_{t-1}$), the current frame ($\mathbf{I}_t$), and the backward optical flow between the current and previous frame ($\mathbf{\hat{O}}_{t-1}^t$). The goal of the Propagation Module is to produce a correction $\mathbf{C}_\mathrm{G}$ to the warped features of the previous frame, $\mathbf{F}^w_{t-1}$. The warping is performed with the refined optical flow $\mathbf{O}_{t-1}^t$. The initial correction tensor $\mathbf{\hat{C}}$ is the gated combination of features that are aware of current appearance, previous geometry, and scene dynamics, \ie $\mathbf{F_I}$, $\mathbf{F}_\mathbf{D}^w$, $\mathbf{F_O}$. The correction tensor is processed with a lightweight encoder to produce the unrefined residuals $\mathbf{C}$, which in turn are made aware of the previous frame features $\mathbf{F}_{t-1}$ and gated based on the flow features $\mathbf{F_O}$. For the sake of simplicity, we present the instances of $\mathbf{C}$, $\mathbf{C}_\mathrm{G}$, $\mathbf{F}_{t-1}$ and $\mathbf{F}^w_{t-1}$ for a single resolution, but these tensors and the block $\mathrm{Residual Refine}$ have four instances, one for each resolution, without sharing weights.}
    \label{fig:method:overview}
\end{figure*}

Video-based data naturally allow the use of prior estimates and the establishment of correspondences between consecutive frames.
However, single-frame monocular depth estimation makes independent per-frame predictions, overlooking these temporal cues.
The inherent temporal coherence in video sequences provides valuable prior information that can be exploited to enhance depth propagation, detailed in \cref{ssec:method:propagation}.
\ourmodel leverages this temporal information by incorporating past depth predictions, deep feature propagation, and refined optical flow in a structured multi-modal framework as depicted in \cref{fig:method:overview}.
The previous depth prediction acts as a geometric prior, ensuring consistency over time.
In the absence of motion, the model should ideally learn an identity transformation from the previous to the current depth prediction, reinforcing temporal stability.
Similarly, deep features from previous frames are propagated to provide additional prior information at the feature level.
Moreover, an additional warping-based 3D self-consistency loss is added to improve consistency over sequences in a bi-directional fashion, as described in \cref{ssec:method:consistency}.

\subsection{Propagation Module}
\label{ssec:method:propagation}

Our Propagation Module is inspired by the video encoding paradigm, where redundancies in video data are compressed by using motion vectors, \ie rough optical flow, and residual coefficients, which capture the difference between the previous frame warped (PFW) and the current frame.
However, unlike video encoding, a video depth estimator does not have access to the current frame output, as the latter is the final product of the entire model.
Therefore, the Propagation Module learns a correction function that refines the PFW output in a way that the corrected version matches the actual current output.
This correction is performed at the deep feature level rather than directly in the output depth space, in order to take advantage of a more expressive feature representation and the pre-trained Decoder.

\PAR{Flow.} Optical flow plays a crucial role in propagation, enabling warping of both previous-frame depth predictions and deep features.
The model encodes flow between consecutive frames using two backbone layers, which take as input a three-channel image: red and green channels encode normalized flow values, while the blue channel captures the luma difference between frames, generating flow features $\mathbf{F_O}$.
The initial optical flow estimate $\mathbf{\hat{O}}^{t}_{t-1}$ is refined using a light-weight two-layer convolutional network, producing a correction term $\mathbf{H}$ that improves the flow accuracy, resulting in a corrected estimate $\mathbf{O}^{t}_{t-1} = \mathbf{\hat{O}}^{t}_{t-1} + \mathbf{H}$.
This refinement step performs denoising and sharpening, particularly beneficial for low-resolution feature space warping.
The initial flow estimate is computed using the CPU-based DIS algorithm~\cite{kroeger1026dis}, but motion vectors or high-quality predicted optical flow can alternatively be used; \ourmodel is agnostic and robust to this initial flow estimation.

\PAR{Fusion and residuals.} While feature warping improves propagation, it is prone to failure in occluded regions or when the optical flow is inaccurate.
To address this, the fusion mechanism incorporates flow features $\mathbf{F_O}$ to guide the selection of reliable propagated information.
RGB and PFW depth features, denoted as $\mathbf{F_I}$ and $\mathbf{F}^w_{\mathbf{D}}$, are first encoded with a backbone layer before being fused using a gated mechanism, which is used to prevent incorrect propagation in regions where $\mathbf{O}^{t}_{t-1}$ is unreliable.
The fused features $\mathbf{\hat{C}}$ are obtained via:
\begin{equation}
\mathbf{\hat{C}} = \mathbf{F_{I}} + \mathbf{G_D} \odot \mathbf{F}^w_{\mathbf{D}} + \mathrm{Lin}_O(\mathbf{F_{O}}),
\end{equation}
where $\mathbf{G_D}$ is the depth gate computed as $\sigma(\mathrm{Lin}_D(\mathbf{F_{O}}))$, and $\mathrm{Lin}$ is a linear layer.
The gating mechanism ensures that erroneous flow does not degrade depth propagation.
The fused features are processed through the remaining blocks of the, now shared, backbone and yield multi-resolution encoder features $\mathcal{C} = \{\mathbf{C}_i\}^4_{i=0}$.
Finally, the warped neck features $\mathbf{F}^w_{i,t-1}$ are corrected using multi-resolution encoder features $\mathcal{C}$ in a residual formulation:
\begin{equation}
\label{eqn:correct}
\mathbf{F}_{i,t} = \mathrm{Conv}(\mathbf{C}_i || \mathbf{F}_{i,t-1}, \mathbf{G_F}) + \mathbf{F}^w_{i,t-1},
\end{equation}
where $\mathbf{G_F}$ is the feature gate controlling the correction process obtained via $\sigma(\mathrm{MLP}_F(\mathbf{F_O}))$, and ``Conv'' is a ResNet Block with gating applied in the bottleneck.
The gating mechanism selectively propagates reliable corrections when needed while filtering out harmful residuals where the PFW depth features are already corrected.
The concatenation of $\mathbf{C}_i || \mathbf{F}_{i,t-1}$ is utilized to make the correction from the Encoder aware of the previous frame features.

\PAR{Keyframe selection.} To ensure efficient propagation, \ourmodel minimizes redundant predictions. If the input remains stable, prior predictions are propagated, while \ourmodel has to re-initialize and predict from scratch when significant changes occur.
In particular, we define a simple re-initialization heuristic based on optical flow via the magnitude of the flow and a warping-based difference metric. Formally, we incur a keyframe if and only if
\begin{equation}
\begin{split}
    \left\lVert w(\mathbf{1}_{H \times W}, \mathbf{\hat{O}}^t_{t-1})\right\rVert_1 &\leq 0.2 \times 0.9^{t} \text{ }\vee \\
    \left\lVert\mathbf{\hat{O}}^t_{t-1}\right\rVert_1 &\geq 0.15 \times 0.9^{t} + 0.1,
\end{split}
\end{equation}
where $\mathbf{1}_{H \times W}$ is a ${H \times W}$ matrix of ones, $w(x, \mathbf{y})$ denotes warping $x$ using flow $\mathbf{y}$, ${\|\mathbf{X}\|}_1 = \frac{1}{HW}\sum_{i,j} {\|x_{ij}\|}_2$, and $t$ is the frame count since the last keyframe.
The decay $0.9^{t}$ accounts for gradual degradation over time, balancing efficiency and robustness for long sequences.

\subsection{Consistency}
\label{ssec:method:consistency}

Maintaining temporal consistency is essential for online and real-time depth estimation.
Ideally, the same 3D point should retain a consistent location across consecutive frames.
However, traditional MDE models operate on independent images; this makes them highly sensitive to small input variations due to the absence of temporal constraints.
To mitigate these issues, \ourmodel introduces a refined consistency loss formulation.
A key limitation of previous methods~\cite{Wang2022fmnet, zhang2019cslstm} is the lack of explicit camera motion compensation.
Depth values propagated through warping reside in different coordinate frames, and without appropriate transformations, their direct comparison is inconsistent.
To ensure equivariance against camera motion, \ourmodel applies the consistency loss on metric radial distance rather than raw depth values. Radial distance remains invariant to rotational transformations, ensuring that consistency is preserved across frames regardless of camera orientation.
To address translational motion, a linear shift is computed by aligning the median-based centers of consecutive 3D point clouds:
\begin{equation}
\begin{split}
\label{eqn:cons}
    \mathcal{L}_{\mathrm{con}}(t-1, t) &= \left\lVert w(\left\lVert\mathbf{P}_{t-1}\right\rVert_2, \mathbf{O}^{t}_{t-1}) - \left\lVert\mathbf{P}_t - \mathbf{t}\right\rVert_2 \right\rVert_1,\\
    \mathbf{t} &= \mathrm{med}(\mathbf{P}_t) - \mathrm{med}(\mathbf{P}_{t-1}),
\end{split}
\end{equation}
where $\mathbf{P} \in \mathbb{R}^{3\times H \times W}$ represents the 3D point map, $\mathbf{O}^{t}_{t-1}$ is the pseudo-ground-truth flow from~\cite{wang2024sea}, and $\mathrm{med}(\cdot)$ computes the median over pixel and dimension-wise elements.
Occlusions and disocclusions are masked out based on a forward-backward flow consistency check as per standard practice.
This formulation enforces a pose-agnostic consistency constraint without requiring explicit extrinsic parameters, enabling robust and efficient depth propagation suitable for practical deployment.
The consistency loss is applicable only for models that infer 3D points directly from RGB inputs, as it is formulated in terms of metric Euclidean distance.
Additionally, the loss is computed bidirectionally, ensuring time-invariant consistency across frames:  $\mathcal{L}_{\mathrm{con}} = \mathcal{L}_{\mathrm{con}}(t-1,t) + \mathcal{L}_{\mathrm{con}}(t,t-1)$.
Moreover, we propose to use it in conjunction with a temporal flip augmentation.
This augmentation helps mitigate the forward-motion bias typically present in casual coherent videos, which would otherwise induce the network to always mimic forward ego-motion even when it is not present.

\subsection{Network Design}
\label{ssec:method:design}

\PAR{Architecture.} The proposed architecture consists of a ``Base Model'', specifically \cite{piccinelli2025unidepthv2}, although the former is adaptable to any metric MDE model, which comprises a ``Base Encoder'' and a ``Base Decoder''.
In addition, \ourmodel involves a propagation network that integrates a multi-modal encoder, residual correction module, and optical flow refinement, as illustrated in \cref{fig:method:overview}.
The multi-modal encoder is a convolutional network, specifically ConvNeXt-Tiny~\cite{liu2022convnext}, with three input branches corresponding to different modalities: RGB, geometric depth, and optical flow.
Each branch extracts dense features $\mathbf{F_X} \in \mathbb{R}^{h \times w \times C \times 4}$, where $(h, w) = (\frac{H}{4}, \frac{W}{4})$, and $\mathbf{X} \in \{\mathbf{I}, \mathbf{D}, \mathbf{O}\}$.
The features are processed through three shared blocks to produce the fused features $\mathbf{C}$, as described in \cref{ssec:method:propagation}.
The processed fused features are multi-scale, producing outputs at four different resolutions, denoted as $\mathcal{C} = \{\mathbf{C}_i\}^3_{i=0}$.
The optical flow refinement module processes $\mathbf{F_O}$ using two convolutional layers interleaved with 2x bilinear upsampling and a leaky ReLU activation function.
The residual module then corrects the neck features at each resolution, $\mathcal{F}_t = \{\mathbf{F}_{i,t}\}^3_{i=0}$, using the multi-modal and multi-resolution features $\mathcal{C}$, as detailed in \eqref{eqn:correct}.
The full Base Model is applied to the first frame, which is treated as a keyframe, to generate the initial neck features, $\mathcal{F}_0$, while the base decoder processes the incoming refined features $\mathcal{F}_t$ for all subsequent frames ($t > 0$) until the next keyframe is incurred as described in \cref{ssec:method:propagation}.
The model outputs the predicted 3D point maps $\mathbf{P}_t \in \mathbb{R}^{3 \times H \times W}$, thanks to intrinsics provided by the Base Model, along with the neck features $\mathcal{F}_t$, which are then propagated to the next frame ($t+1$).

\PAR{Optimization.} The optimization strategy comprises five distinct loss functions targeting three main objectives: output accuracy, flow refinement, and consistency.
Depth predictions are optimized using the $\mathrm{SI}_{\log}(\mathbf{D}^*, \mathbf{D})$ loss from~\cite{Eigen2014}, where $\mathbf{D}^*$ denotes ground-truth depth, and the $\mathrm{L_{1,SSI}}(\mathbf{D}^*, \mathbf{D})$ loss from~\cite{ranftl2020midas}, computed over the entire video clip rather than per image.
When GT depth is unavailable, the supervision is derived from the ``Base Model'' predictions.
The stability and accuracy of depth predictions depend on ensuring that neck features remain sharp and do not degrade due to warping.
Therefore, the corrected neck features are supervised by aligning them to the per-frame Base Model features ($\{\mathbf{F}^*_{i,t}\}^3_{i=0}$) using an $L_1$ loss:
$\mathcal{L}_{\mathrm{F}}(\mathcal{F}^*_t, \mathcal{F}_t) = \sum^4_{i=1} \left( \frac{1}{C}\sum_{c=1}^C{\| \mathbf{F}^*_{i,t,c} - \mathbf{F}_{i,t,c}\|}_1\right)$.
The refined optical flow $\mathbf{O}$ is supervised using pseudo-GT backward flow produced by SEA-RAFT~\cite{wang2024sea} with an $L_1$ loss.
Finally, the consistency between consecutive frames is enforced through the proposed bidirectional consistency loss $\mathcal{L}_{\mathrm{con}}$ described in \cref{ssec:method:consistency}.
This formulation ensures that depth predictions remain stable over time while enabling accurate depth propagation across video sequences.
The final loss is the sum.

\begin{figure}[t]
    \centering
    \footnotesize
    \includegraphics[width=1.0\linewidth]{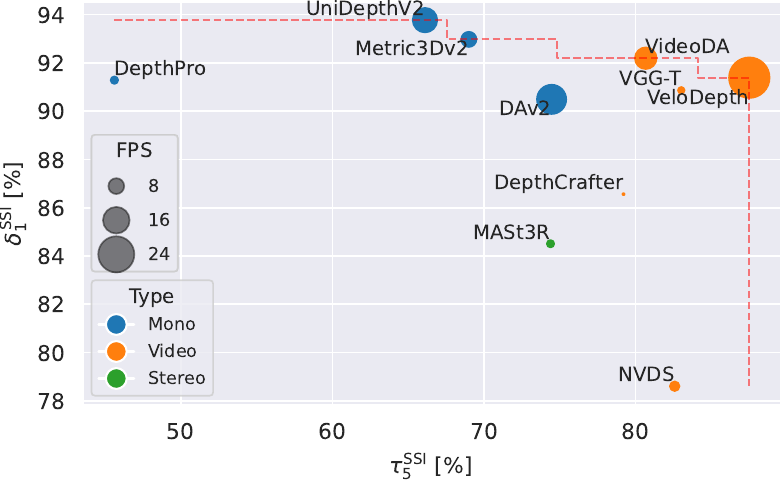}
    \vspace{-10pt}
    \caption{\textbf{Pareto optimal frontier} is evaluated in terms of combined accuracy ($\delta^{\mathrm{SSI}}_1$) and consistency ($\tau^{\mathrm{SSI}}_5$).
    Disk areas correspond to inference efficiency (FPS); the larger the area, the faster.
    \ourmodel strikes a positive tradeoff \wrt its Base Model (\ie UniDepthV2), achieving a substantial improvement in consistency for a minor drop in accuracy.}
    \label{fig:method:pareto}
\end{figure}

\section{Experiments}
\label{sec:experiments}

\begin{figure*}[ht]
    \renewcommand{\arraystretch}{1}
    \centering
    \small
    \begin{tabular}{ccc|ccc|c}
        \multirow{1}{*}[2.9em]{\rotatebox[origin=c]{90}{Bonn}}
        & \includegraphics[width=0.15\linewidth]{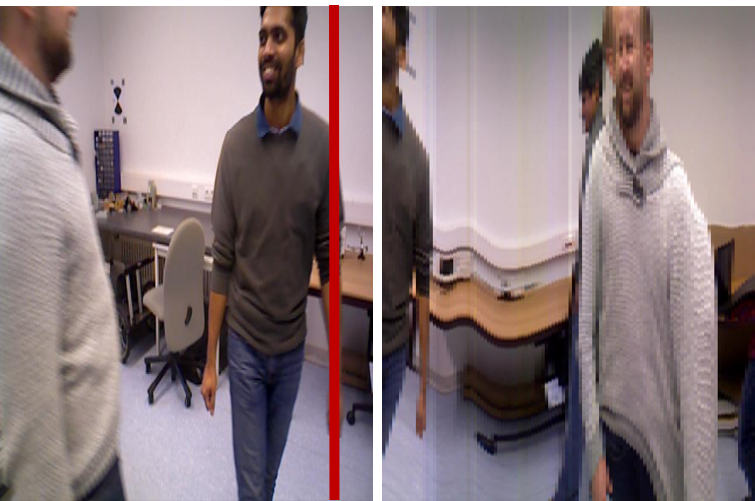}
        & \includegraphics[width=0.15\linewidth]{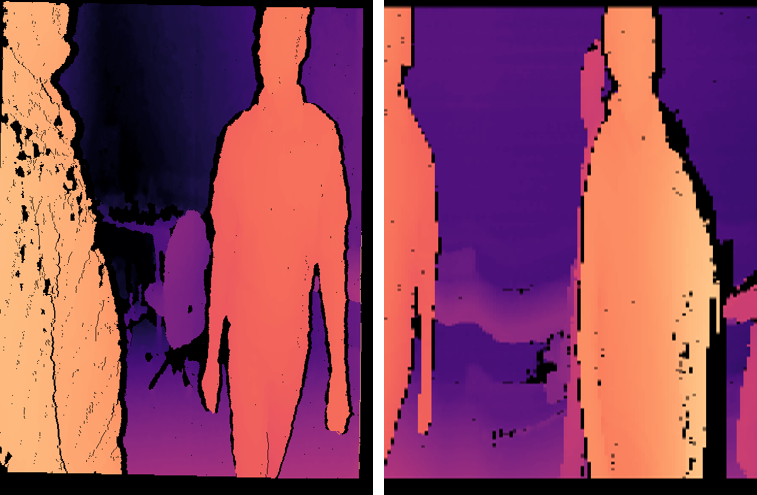}
        & \includegraphics[width=0.15\linewidth]{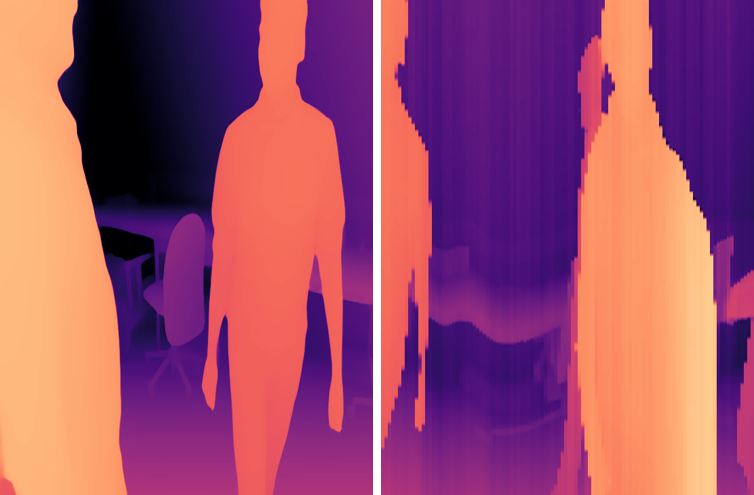}
        & \includegraphics[width=0.15\linewidth]{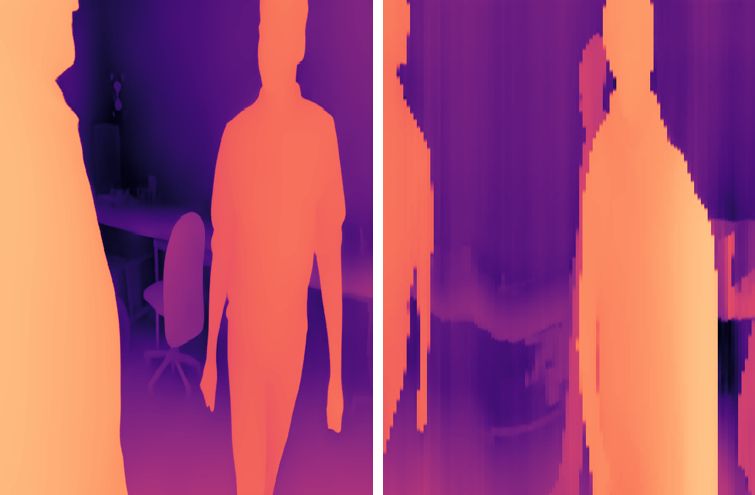}
        & \includegraphics[width=0.15\linewidth]{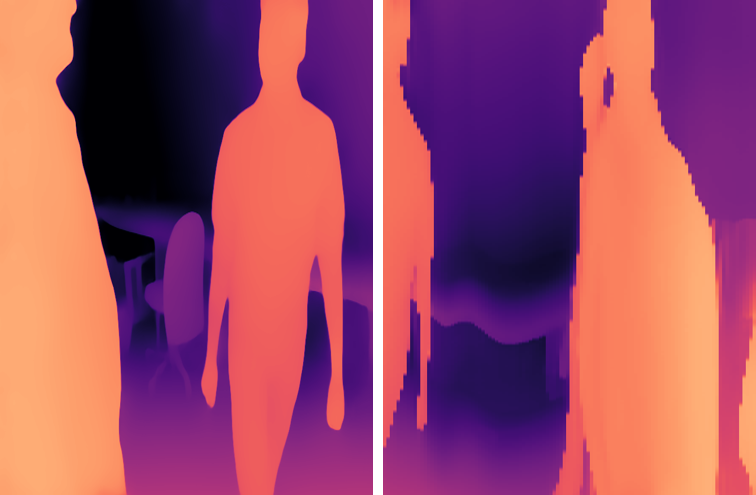}
        & \includegraphics[width=0.025\linewidth]{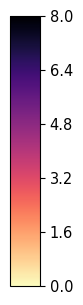}\\

        \multirow{1}{*}[3.0em]{\rotatebox[origin=c]{90}{TUM}}
        & \includegraphics[width=0.15\linewidth]{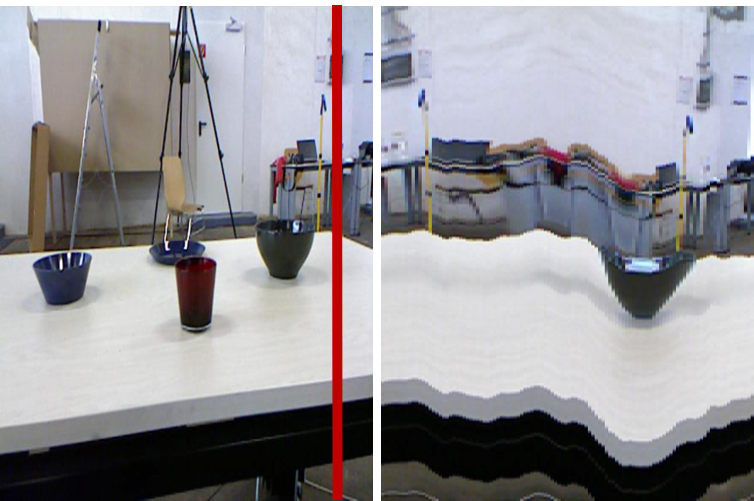}
        & \includegraphics[width=0.15\linewidth]{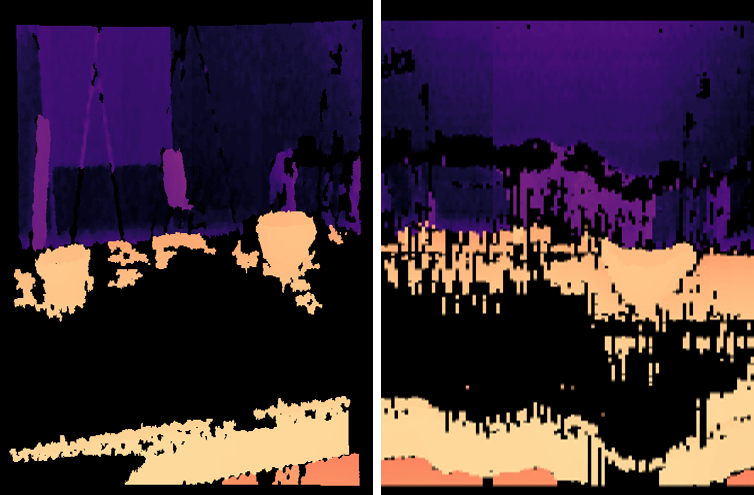}
        & \includegraphics[width=0.15\linewidth]{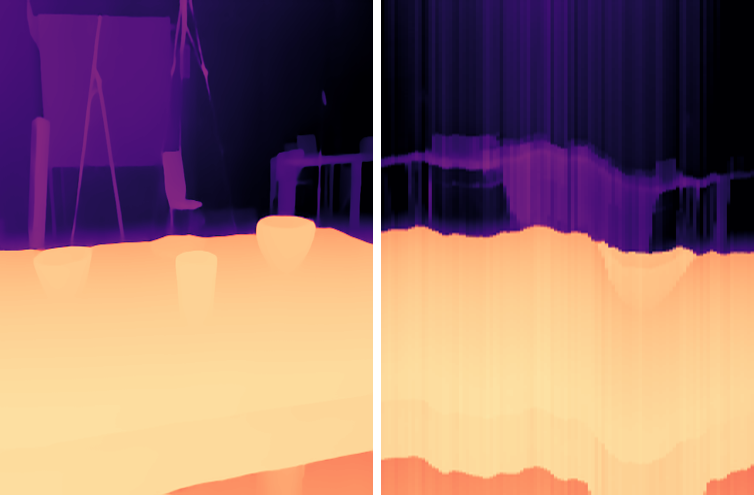}
        & \includegraphics[width=0.15\linewidth]{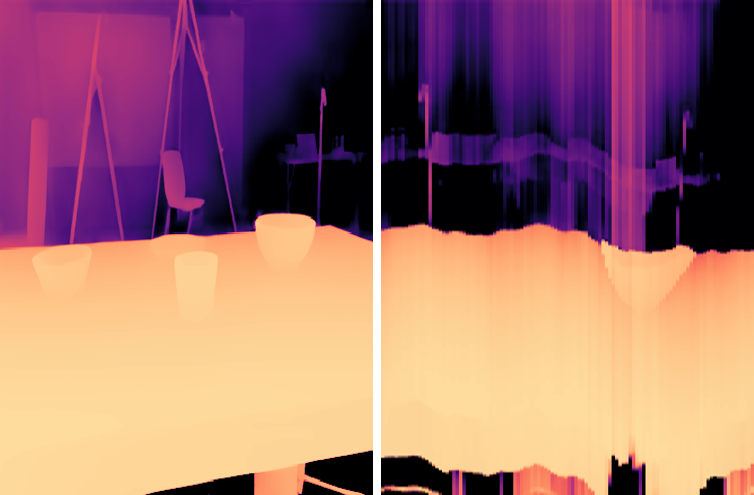}
        & \includegraphics[width=0.15\linewidth]{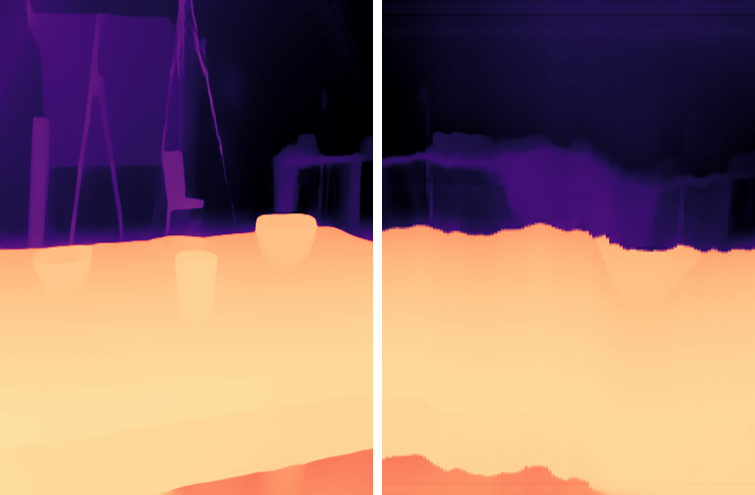}
        & \includegraphics[width=0.025\linewidth]{figures/qualitative/colorbar_8.png}\\

        \multirow{1}{*}[3.0em]{\rotatebox[origin=c]{90}{Sintel}}
        & \includegraphics[width=0.15\linewidth]{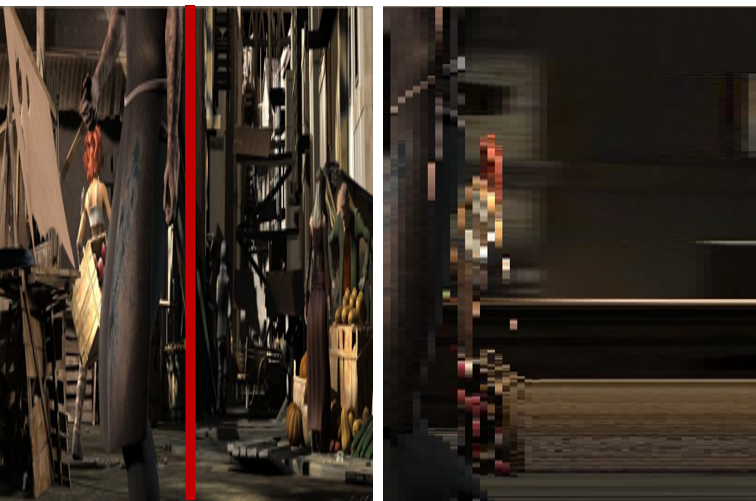}
        & \includegraphics[width=0.15\linewidth]{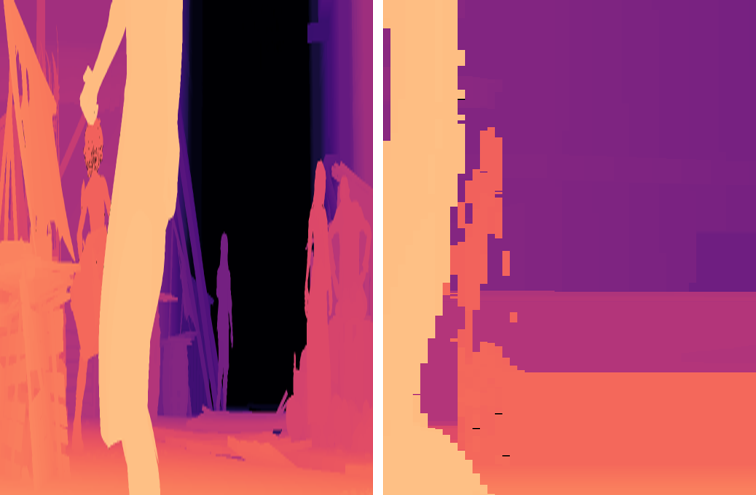}
        & \includegraphics[width=0.15\linewidth]{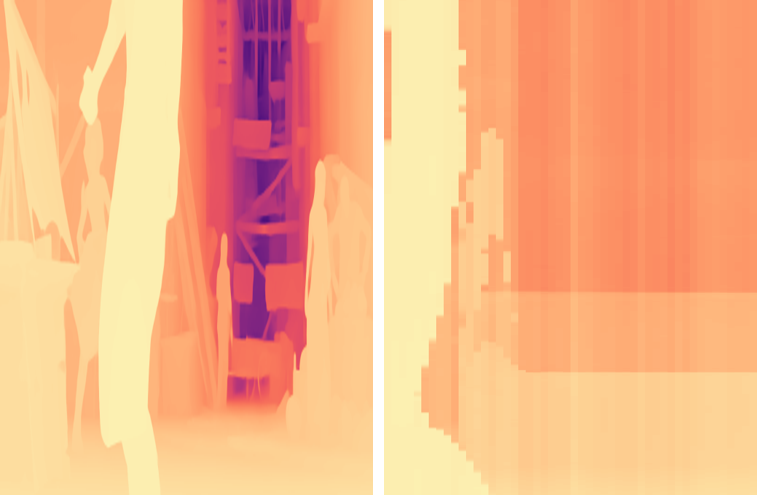}
        & \includegraphics[width=0.15\linewidth]{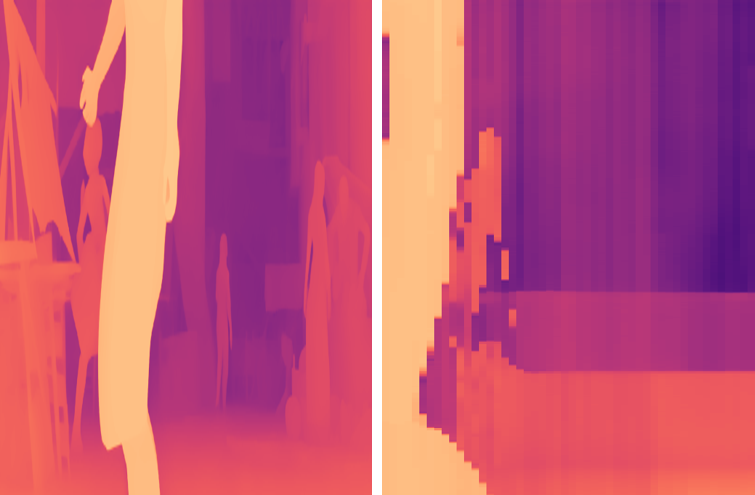}
        & \includegraphics[width=0.15\linewidth]{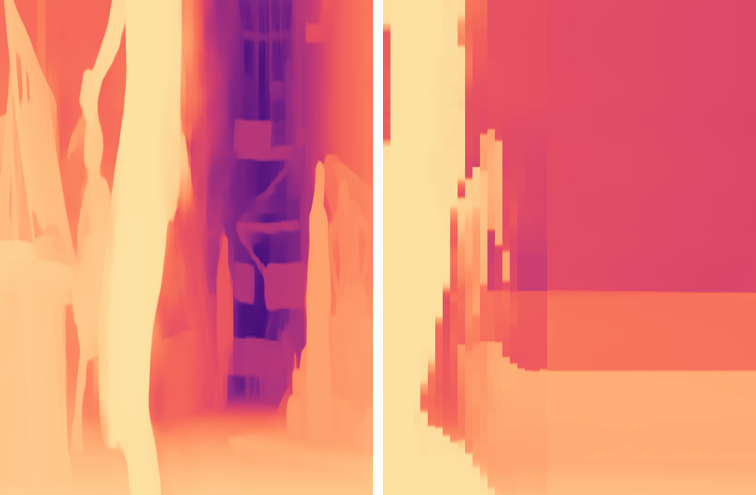}
        & \includegraphics[width=0.025\linewidth]{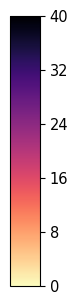}\\

        \multirow{1}{*}[3.5em]{\rotatebox[origin=c]{90}{ScanNet}}
        & \includegraphics[width=0.15\linewidth]{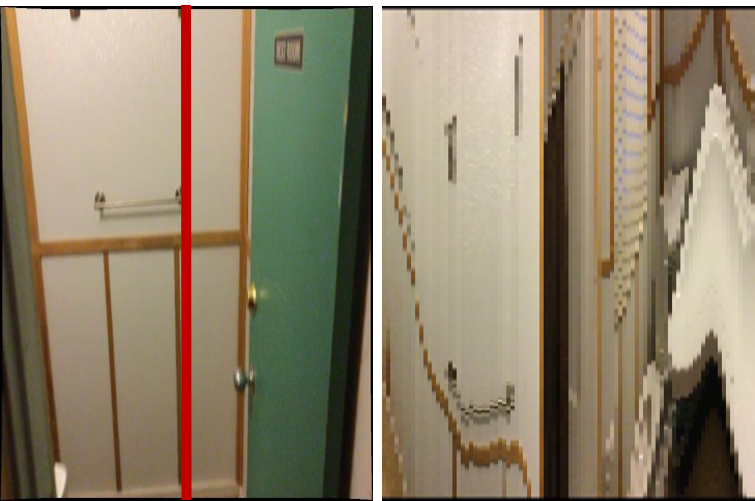}
        & \includegraphics[width=0.15\linewidth]{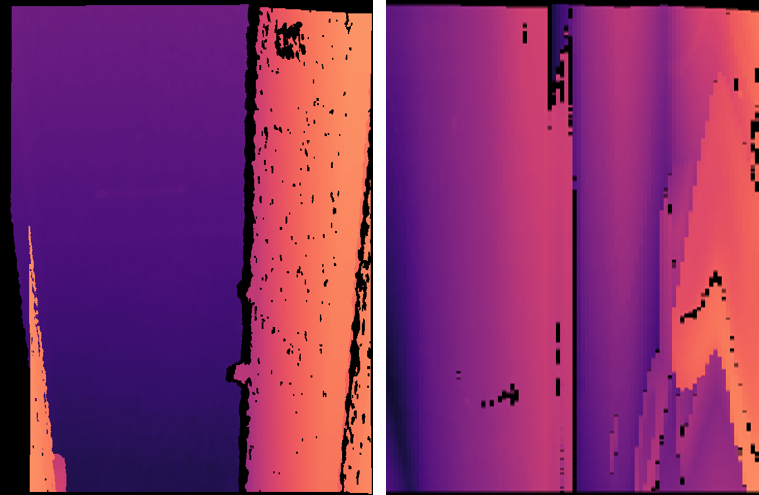}
        & \includegraphics[width=0.15\linewidth]{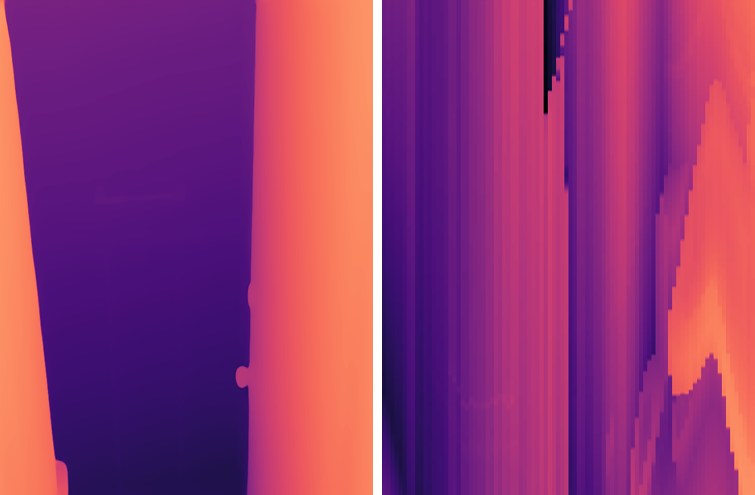}
        & \includegraphics[width=0.15\linewidth]{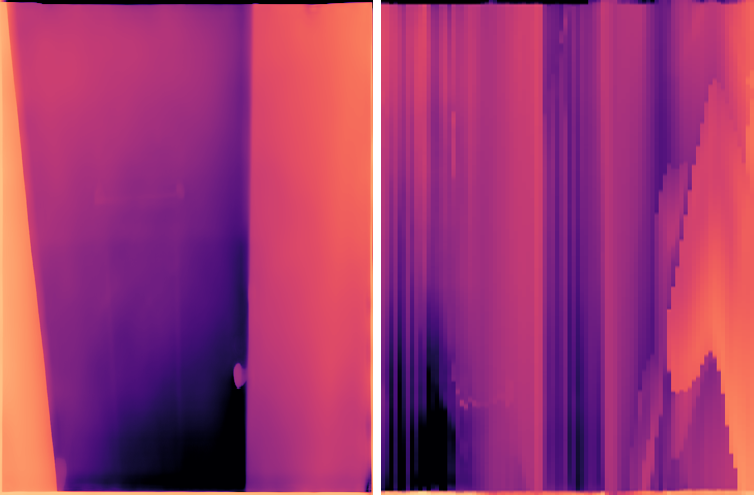}
        & \includegraphics[width=0.15\linewidth]{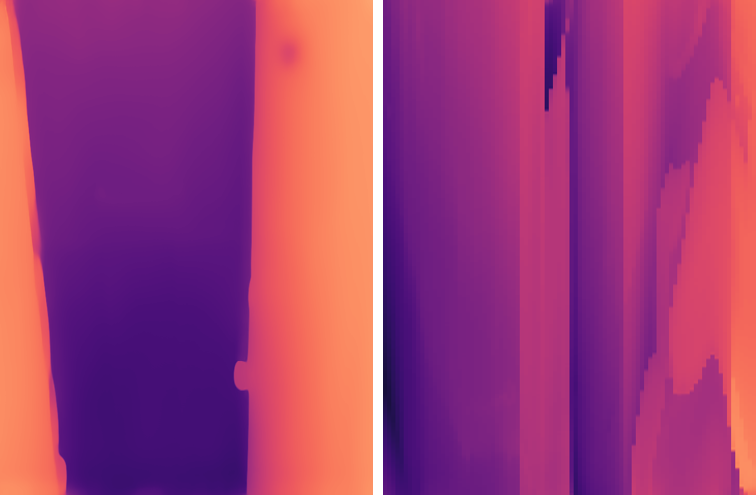}
        & \includegraphics[width=0.025\linewidth]{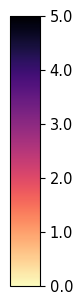}\\
    
        & RGB & GT & UniK3D~\cite{piccinelli2025unik3d} & DepthCrafter\textsuperscript{\dag}~\cite{hu2024crafter} & \ourmodel & Depth \\
    \end{tabular}
    \vspace{-10pt}
    \caption{\textbf{Zero-shot qualitative results.} Each row corresponds to one test video sample from one domain. Each block shows the 6th frame and the video slices corresponding to the red line x-location in the first column. UniDepthV2 and \ourmodel outputs are inherently metric. No post-processing is applied. The last column represents the depth values \wrt ``magma'' colormap. (\dag): affine transformed to match GT. Best viewed on a screen and zoomed in.}
    \label{fig:results:main_vis}
\end{figure*}

\subsection{Experimental Setup}
\label{ssec:experiments:setup}

\noindent\textbf{Datasets.} The training dataset accounts for two different sources, in-the-wild without GT and depth datasets.
The former is composed by Kinetics-700~\cite{smaira2020kinetics700}, Moments-in-Time~\cite{monfort2019moments}, and SAv2~\cite{ravi2024sam}, while the latter by TartanAir~\cite{wang2020tartanair}, Wild-RGBD~\cite{xia2024wildrgbd}, HabitatMatterport3D~\cite{ramakrishnan2021habitat}, PointOdyssey~\cite{zheng2023pointodyssey} and Waymo~\cite{sun2020waymo}.
More details are given in the supplement.
We evaluate the generalizability of models by testing them on 4 datasets not seen during training, in particular, ScanNet~\cite{dai2017scannet}, 
Sintel~\cite{Butler2012sintel}, Bonn-RGBD~\cite{palazzolo2019bonn}, and TUM-RGBD~\cite{sturm12tum}.

\PAR{Implementation details.} \ourmodel is implemented in PyTorch~\cite{pytorch} and CUDA~\cite{nickolls2008cuda}.
Training uses the AdamW~\cite{Loshchilov2017adamw} optimizer ($\beta_1=0.9$, $\beta_2=0.999$) with an initial learning rate of $1\times{}10^{-4}$.
A cosine annealing scheduler reduces the learning rate to one-tenth after 30\% of total training iterations.
We run 150k optimization iterations with 256 total images per iteration.
The dataset sampling procedure follows a weighted sampler, where each dataset is weighted by its number of scene.
We employ curriculum learning to progressively increase sequence length from 2 to 20 frames, using a linear schedule between 50k and 150k iterations.
The idea behind curriculum learning is a progressive increase in sequence complexity, which stabilizes training when handling long video sequences.
Since a single GPU can accommodate only 10 non-keyframe frames per iteration, initial frames of longer sequences are processed in ``no grad'' context.
Our augmentations are both geometric and photometric, \ie random resizing and cropping for the former type, and brightness, gamma, saturation, and hue shift for the latter.
In addition, we employ temporal augmentation which flips the ordering of the frames in each batch with 50\% probability.
The Base Encoder and Decoder are frozen and initialized with UniK3D~\cite{piccinelli2025unik3d} weights.
We randomly sample the image ratio per batch between 2:1 and 1:2 and between 0.2 and 0.4 Megapixel (MP).
The training time amounts to 6 days on 8 NVIDIA RTX 4090.
For the ablations, we run 80k training steps with the training pipeline as for the main experiments compressed from 150k to 80k steps.

\PAR{Evaluation details.} We evaluate on ScanNet following protocol from~\cite{ke2024video} and on Bonn-RGBD and TUM-RGBD following~\cite{hu2024crafter}, while for Sintel all sequences are tested.
Depth accuracy and consistency are assessed using $\delta_1$ and $\tau_5$ metrics, respectively.
$\delta_1$ measures the percentage of pixels whose predicted depth is within 25\% of the GT depth.
$\tau_5$ measures consistency across frames by warping depth from $t-1 \rightarrow t$ using optical flow, applying ego-motion correction, and considering a pixel inlier when the difference is within 5\% of the depth at $t$.
This metric extends the accuracy evaluation used in OPW~\cite{Wang2022fmnet} and TCM~\cite{zhang2019cslstm}, incorporating additional ego-motion compensation.
Optical flow is either sourced from~\cite{wang2024sea} or provided by the dataset itself.
When per-frame depth predictions are rescaled to match GT depth for $\delta_1$ or $\tau_5$, we denote them as $\delta^{\mathrm{SSI}}_1$ and $\tau^{\mathrm{SSI}}_5$.
This rescaling enables fair comparisons with non-metric models while ignoring global scale inconsistencies in $\tau_5$.
GPU inference speed is measured on an NVIDIA RTX 3090 using synchronized timers.
CPU inference speed is evaluated on an M1 Pro chip utilizing the MPS backend, as this setup closely approximates modern mobile processors such as A19 chip while keeping the testing simpler.
Inference speed is measured over a 60-frame sequence and averaged per frame on 0.5 MP images.
Both GPU and CPU benchmarks employ mixed precision.
For the ablations, we evaluate \ourmodel by running it on the first 32 frames of each sequence and initializing on the first frame with the Base Model.
All methods are evaluated in an online fashion: at frame $t$, each model has access to only frames $\le t$.
Direct comparisons to models operating offline would be misleading, as the latter exploit future information, which is not possible in causal settings.
We provide offline evaluation in the supplements.

\subsection{Comparison with The State of The Art}
\label{ssec:experiments:sota}

\begin{table*}[ht]
\centering
\caption{\textbf{Comparison on zero-shot evaluation.} All methods are evaluated in an online fashion. The ``Type'' column indicates the original task tackled, \textbf{M}: monocular, \textbf{S}: stereo, \textbf{V}: video or multi-view. Profiling is run on 60 frames of 0.5MP averaged per frame. $\mathrm{FPS_{GPU}}$ is measured on an RTX 3090 and $\mathrm{FPS_{CPU}}$ on an M1 chip, both with half precision. \ddag: camera GT at inference time.}
\vspace{-10pt}
\label{tab:results:comparison}
\resizebox{\linewidth}{!}{%
\begin{tabular}{lccccccccccccccccc}
\toprule
\multirow{2}{*}{\textbf{Method}} & \multirow{2}{*}{\textbf{Type}} & \multicolumn{2}{c}{TUM-RGBD} & \multicolumn{2}{c}{ScanNet} & \multicolumn{2}{c}{Sintel} & \multicolumn{2}{c}{Bonn-RGBD} & \multicolumn{2}{c}{\textbf{Aggregate}} & \multicolumn{4}{c}{\textbf{Efficiency}} \\
\cmidrule(lr){3-4} \cmidrule(lr){5-6} \cmidrule(lr){7-8} \cmidrule(lr){9-10} \cmidrule(lr){11-12} \cmidrule(lr){13-16}
 &  & $\mathrm{\delta^{SSI}_1}\uparrow$ & $\mathrm{\tau^{SSI}_{5}}\uparrow$ & $\mathrm{\delta^{SSI}_1}\uparrow$ & $\mathrm{\tau^{SSI}_{5}}\uparrow$ & $\mathrm{\delta^{SSI}_1}\uparrow$ & $\mathrm{\tau^{SSI}_{5}}\uparrow$ & $\mathrm{\delta^{SSI}_1}\uparrow$ & $\mathrm{\tau^{SSI}_{5}}\uparrow$ & $\mathrm{\delta^{SSI}_1}\uparrow$ & $\mathrm{\tau^{SSI}_{5}}\uparrow$ & $\mathrm{FPS_{GPU}}\uparrow$ & $\mathrm{FPS_{CPU}}\uparrow$ & $\mathrm{Params [M]}\downarrow$ & $\mathrm{FLOP [T]}\downarrow$ \\
\midrule
DAv2~\cite{yang2024da2} & M & $94.0$ & $79.9$ & $98.1$ & $72.8$ & $73.0$ & $55.0$ & $98.5$ & $95.1$ & $90.5$ & $74.5$ & $16.9$ & $1.1$ & $335.3$ & $2.0$ \\
Metric3Dv2\textsuperscript{\ddag}~\cite{hu2024metric3dv2} & M & $\scnd{96.1}$ & $74.0$ & $\scnd{99.0}$ & $81.3$ & $77.2$ & $31.3$ & $\best{99.1}$ & $86.0$ & $93.0$ & $69.0$ & $7.1$ & $0.7$ & $411.9$ & $3.5$ \\
DepthPro~\cite{bochkovskii2024depthpro} & M & $94.3$ & $58.2$ & $97.2$ & $43.9$ & $73.7$ & $17.6$ & $\scnd{99.0}$ & $63.2$ & $91.3$ & $45.6$ & $2.8$ & $0.2$ & $952.0$ & $4.8$ \\
UniDepthV2~\cite{piccinelli2025unidepthv2} & M & $\best{96.6}$ & $71.0$ & $98.5$ & $74.8$ & $\best{80.7}$ & $34.5$ & $\scnd{99.0}$ & $82.0$ & $\best{93.8}$ & $66.1$ & $13.4$ & $1.0$ & $353.8$ & $2.2$ \\
UniK3D~\cite{piccinelli2025unik3d} & M & $\best{96.6}$ & $69.4$ & $98.5$ & $76.1$ & $\scnd{80.5}$ & $32.2$ & $\scnd{99.0}$ & $79.7$ & $\scnd{93.6}$ & $64.4$ & $12.1$ & $1.0$ & $375.3$ & $2.6$ \\
MASt3R~\cite{leroy2024master} & S & $88.1$ & $77.4$ & $96.7$ & $77.9$ & $59.3$ & $55.6$ & $93.9$ & $86.6$ & $84.5$ & $74.4$ & $2.7$ & $0.7$ & $688.6$ & $3.2$ \\
CS-LSTM~\cite{zhang2019cslstm} & V & $24.7$ & $81.6$ & $26.0$ & $59.7$ & $8.0$ & $\scnd{71.1}$ & $27.7$ & $73.0$ & $20.8$ & $69.0$ & $\scnd{25.3}$ & $\scnd{2.5}$ & $15.0$ & $\best{0.4}$ \\
NVDS~\cite{wang2023nvds} & V & $76.6$ & $83.9$ & $85.1$ & $\scnd{88.7}$ & $67.6$ & $65.8$ & $88.0$ & $\best{97.3}$ & $78.6$ & $82.6$ & $3.9$ & $0.5$ & $432.9$ & $2.2$ \\
ChronoDepth~\cite{shao2024chrono} & V & $58.5$ & $76.6$ & $74.1$ & $72.4$ & $26.5$ & $34.2$ & $61.8$ & $68.8$ & $55.2$ & $63.0$ & $0.3$ & \textcolor{gray}{OOM} & $1522.3$ & $19.6$ \\
DepthCrafter~\cite{hu2024crafter} & V & $83.5$ & $80.0$ & $94.4$ & $83.2$ & $74.3$ & $64.9$ & $98.7$ & $96.3$ & $86.6$ & $79.2$ & $0.2$ & \textcolor{gray}{OOM} & $1524.6$ & $27.1$ \\
VideoDA~\cite{chen2025vda} & V & $94.9$ & $80.1$ & $98.1$ & $87.5$ & $76.8$ & $64.2$ & $99.0$ & $96.0$ & $92.2$ & $80.7$ & $11.3$ & \textcolor{gray}{OOM} & $384.4$ & $3.9$ \\
VGG-T~\cite{wang2025vggt} & V & $90.2$ & $\scnd{84.4}$ & $\best{99.3}$ & $86.5$ & $74.1$ & $68.9$ & $\scnd{99.0}$ & $\scnd{96.6}$ & $90.9$ & $\scnd{83.0}$ & $2.3$ & \textcolor{gray}{OOM} & $1261.0$ & $8.8$ \\
\midrule
\ourmodel & V & $94.8$ & $\best{89.1}$ & $96.2$ & $\best{89.1}$ & $76.1$ & $\best{75.4}$ & $98.4$ & $96.3$ & $91.4$ & $\best{87.5}$ & $\best{26.2}$ & $\best{2.7}$ & $409.4$ & $\scnd{0.7}$ \\
\bottomrule
\end{tabular}%
}
\end{table*}

\Cref{tab:results:comparison} presents a comprehensive evaluation of \ourmodel against state-of-the-art monocular, stereo, and video depth estimation methods across four distinct domains.
It is worth noting that we report mainly scale- and shift-invariant metrics to increase the extensiveness of our comparison, but \ourmodel outputs metric predictions, which are evaluated more extensively in the supplements.
Our method clearly demonstrates superior temporal consistency and computational efficiency compared to all competitors.
In particular, when compared to a model with a similar runtime, such as \cite{zhang2019cslstm}, \ourmodel achieves significantly higher accuracy (+70.6\%) and consistency (+18.5\%).
Furthermore, compared to the closest competitor in terms of consistency, \cite{wang2023nvds}, our approach not only improves accuracy by 12.8\%, but also provides a $6.7\times$ improvement in inference speed.
However, \ourmodel can produce metric output, in contrast to most video-based methods, and in the metric-case, it ranks 1\textsuperscript{st} and 2\textsuperscript{nd} for consistency and accuracy, respectively. 

While monocular depth estimators generally yield higher absolute accuracy, this accuracy typically comes at the expense of temporal consistency. For instance, \ourmodel notably surpasses the consistency of its monocular base model (UniK3D) by 13.3\% and 21.4\% for metric and affine-invariant evaluation, respectively, highlighting the strength of our propagation-based approach.
Moreover, as illustrated in \cref{fig:results:main_vis}, despite being a metric depth estimator susceptible to global scale jitter, unlike relative depth estimators, ourmodel still maintains remarkably high consistency.
Traditional monocular and repurposed-online depth methods exhibit substantial frame-to-frame jitter as color jumps, indicative of inconsistent predictions, whereas \ourmodel effectively mitigates this issue through its feature propagation mechanism.
It is important to note that \ourmodel inherits occasional inconsistencies from keyframe predictions produced by the monocular base model, especially when significant scene changes trigger the computation of a new keyframe.
Despite this, the propagated intermediate predictions remain highly stable.

Finally, as depicted in \cref{fig:method:pareto}, \ourmodel establishes a Pareto-optimal frontier, clearly demonstrating the best available trade-off between consistency, accuracy, and computational efficiency among current depth estimation methods.

\subsection{Ablations}
\label{ssec:experiments:ablations}

\begin{figure}[t]
    \centering
    \footnotesize
    \includegraphics[width=0.9\linewidth]{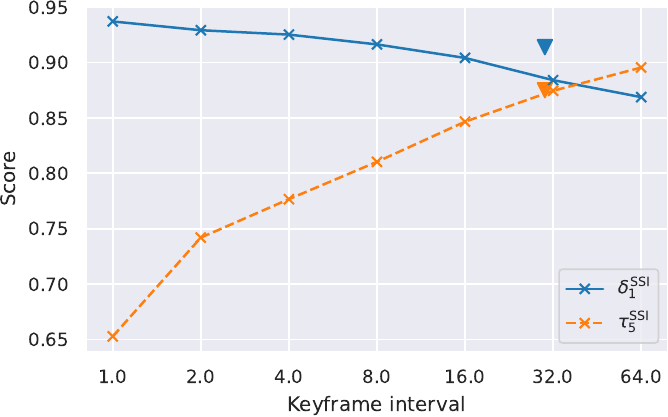}
    \vspace{-10pt}
    \caption{\textbf{Keyframe interval impact} is evaluated in accuracy ($\delta^{\mathrm{SSI}}_1$) and consistency ($\tau^{\mathrm{SSI}}_5$).
    Plot lines refer to fixed keyframe intervals on the x-axis.
    $\blacktriangledown$ refers to our keyframe selection mechanism, accounting for an average keyframe interval of 30.}
    \label{fig:method:kf_selection}
\end{figure}

We conduct an extensive ablation study to evaluate the impact of key architectural and optimization components.
This includes analyzing input modalities in \Cref{tab:ablations:inputs}, gating mechanisms in \Cref{tab:ablations:gating}, loss functions in \Cref{tab:ablations:loss}, the choice of optical flow in \Cref{tab:ablations:flow}, and the proposed inductive biases in the Propagation Module in \Cref{tab:ablations:propagation} and keyframe selection in \cref{fig:method:kf_selection}.
Each table underlines a row, which corresponds to the (partial) configuration used in the final \ourmodel.

\begin{table}[t]
\centering
\caption{\textbf{Input modalities.} $\mathbf{D}^w_{t-1}$ indicates the previous frame warped depth and $\mathbf{\hat{O}}^{t}_{t-1}$ the initial optical flow. Current RGB is always used as input.}
\vspace{-10pt}
\label{tab:ablations:inputs}
\begin{tabular}{ccc|cccc}
    \toprule
     & $\mathbf{D}^{w}_{t-1}$ & $\mathbf{\hat{O}}^{t}_{t-1}$ & $\mathrm{\delta_1}\uparrow$ & $\mathrm{\tau_5}\uparrow$ & $\mathrm{\delta^{SSI}_1}\uparrow$ & $\mathrm{\tau^{SSI}_5}\uparrow$ \\
    \midrule
    1 & \xmark & \xmark & $63.1$ & $78.0$ & $76.8$ & $81.1$ \\
    2 & \cmark & \xmark & $62.3$ & $88.3$ & $78.4$ & $90.6$ \\ 
    3 & \xmark & \cmark & $69.3$ & $89.7$ & $82.6$ & $90.3$ \\
    \underline{4} & \cmark & \cmark & $71.3$ & $91.8$ & $82.9$ & $92.4$ \\
    \bottomrule
    \end{tabular}%
    
\end{table}
\PAR{Input Modalities.} The results in \Cref{tab:ablations:inputs} highlight the role of different input modalities in \ourmodel.
Adding the PFW depth prediction (row 2) significantly enhances depth consistency, demonstrating the importance of propagating prior depth estimates.
However, this addition does not directly improve depth accuracy, suggesting that the network primarily learns to preserve existing structures, \ie zero residuals $\mathbf{C}_\mathrm{G}$ rather than actively refining depth estimates.
Integrating optical flow further improves both consistency and accuracy by allowing the model to identify incorrect PFW features.
This enables targeted feature corrections while maintaining stability by setting the residual to zero in regions where depth estimates are already reliable.
The best performance (row 4) is achieved when both PFW depth and optical flow are included, as \ourmodel gains a comprehensive understanding of prior depth information, its motion dynamics, and where to trust these estimates.
The RGB image is always included as a reference modality in all experiments.

\begin{table}[t]
\centering
\caption{\textbf{Flow-based gating.} $\mathbf{F}^{w}_{t-1}$ corresponds to previous frame warped decoder features, $\mathbf{F}^{w}_\mathbf{D}$ to depth features, and $\mathbf{F_I}$ image features after the first respective layers. $\sigma$ indicates if the element-wise sigmoid-based gating is applied. ``$1 - \sigma$'' represents inverse gating \wrt $\mathbf{F}^{w}_\mathbf{D}$ one.}
\vspace{-10pt}
\label{tab:ablations:gating}
\resizebox{\linewidth}{!}{%
\begin{tabular}{cccc|cccc}
    \toprule
     & $\mathbf{F}^{w}_{t-1}$ & $\mathbf{F}^{w}_\mathbf{D}$ & $\mathbf{F_I}$ & $\mathrm{\delta_1}\uparrow$ & $\mathrm{\tau_5}\uparrow$ & $\mathrm{\delta^{SSI}_1}\uparrow$ & $\mathrm{\tau^{SSI}_5}\uparrow$ \\
    \midrule
    1 & \xmark & \xmark & \xmark         & $71.3$ & $91.8$ & $82.9$ & $92.4$ \\
    2 & $\sigma$ & \xmark & \xmark       & $73.1$ & $93.0$ & $84.3$ & $93.9$ \\
    \underline{3} & $\sigma$ & $\sigma$ & \xmark & $74.1$ & $92.7$ & $84.7$ & $93.5$ \\ 
    4 & $\sigma$ & $\sigma$ & $1-\sigma$ & $72.8$ & $91.7$ & $83.9$ & $92.5$ \\
    \bottomrule
    \end{tabular}%
}
\end{table}
\PAR{Gating Mechanisms.} \Cref{tab:ablations:gating} presents an analysis of the gating mechanisms applied at different stages in the network.
The most significant impact is observed when gating is applied to PFW neck features (row 2), as it directly regulates whether residual corrections from the Propagation Module influence the next frame.
This prevents the network from being implicitly biased to predicting always zeros, since for large parts of images correction is typically null and supervised to be null.
Rows 3 and 4 evaluate gating during modality fusion, where gating depth slightly improves accuracy.
However, this effect is partially redundant, as the PFW feature gating (row 2) already ensures that residuals are only applied when necessary, effectively preventing unnecessary corrections.
In row 4, an inverse gate is introduced on RGB features, enforcing a convex combination of depth and RGB information.
However, this leads to a detrimental effect on performance. We speculate that this occurs because RGB features are never warped, meaning that applying a gating function to them results in information loss rather than selective refinement.

\begin{table}[t]
\centering
\caption{\textbf{Optimization.} $\mathrm{Flip}$ refers to using the temporal flipping augmentation.  $\mathcal{L}_\mathrm{con}$ and $\mathrm{out}_{\mathcal{L}_\mathrm{con}}$ indicate if the proposed consistency loss is employed and on which output, respectively, with $\mathbf{D}$ referring to depth and $\mathbf{R}$ to euclidean distance.}
\vspace{-10pt}
\label{tab:ablations:loss}
\resizebox{\linewidth}{!}{%
\begin{tabular}{cccc|cccc}
    \toprule
     & Flip & $\mathcal{L}_\mathrm{con}$ & $\mathrm{out}_{\mathcal{L}_\mathrm{con}}$ & $\mathrm{\delta_1}\uparrow$ & $\mathrm{\tau_5}\uparrow$ & $\mathrm{\delta^{SSI}_1}\uparrow$ & $\mathrm{\tau^{SSI}_5}\uparrow$  \\
    \midrule
    1 & \xmark & \xmark & $\mathrm{n/a}$ & $68.3$ & $89.9$ & $81.6$ & $90.4$ \\
    2 & \cmark & \xmark & $\mathrm{n/a}$ & $72.5$ & $90.6$ & $83.8$ & $92.2$ \\
    3 & \cmark & \cmark & $\mathbf{D}$ & $71.5$ & $88.8$ & $82.6$ & $92.1$ \\
    \underline{4} & \cmark & \cmark & $\mathbf{R}$ & $74.1$ & $92.7$ & $84.7$ & $93.5$ \\
    \bottomrule
    \end{tabular}%
}
\end{table}
\PAR{Loss.} \Cref{tab:ablations:loss} presents the ablation results on the training pipeline, focusing on flip augmentation (row 2) and the proposed loss functions (rows 3 and 4).
The results indicate that flip augmentation enhances accuracy by mitigating the forward motion mimicking bias, as discussed in \cref{ssec:method:consistency}.
The proposed consistency loss, introduced in \cref{ssec:method:consistency}, significantly improves both depth consistency and accuracy.
By enforcing similarity between matching locations in consecutive frames, up to a translation, the loss provides an additional supervision signal that reinforces temporal stability.
Conversely, applying the consistency loss directly to depth values instead of Euclidean distances leads to a performance drop, as shown in row 3.
This result suggests that enforcing consistency in depth space alone introduces incorrect supervision signals, leading to inconsistencies in depth predictions.

\begin{table}[t]
\centering
\caption{\textbf{Flow.} $\mathbf{O}^{t}_{t-1}$ is the flow used to perform warping. $\mathrm{MV}$ refers to using MPEG-4 motion vectors, $\mathrm{DIS}$ utilizes~\cite{kroeger1026dis} and $\mathrm{RAFT}$~\cite{wang2024sea}. The subscript $\mathrm{R}$ stands for usage of the corresponding optical flow refined via $\mathrm{FlowRefine}$.}
\vspace{-10pt}
\label{tab:ablations:flow}
\begin{tabular}{cc|cccc}
    \toprule
     & $\mathbf{O}^{t}_{t-1}$ & $\mathrm{\delta_1}\uparrow$ & $\mathrm{\tau_5}\uparrow$ & $\mathrm{\delta^{SSI}_1}\uparrow$ & $\mathrm{\tau^{SSI}_5}\uparrow$  \\
    \midrule
    1 & $\mathrm{MV}$ & $69.9$ & $89.9$ & $79.4$ & $90.2$\\
    2 & $\mathrm{DIS}$ & $70.2$ & $90.8$ & $81.8$ & $91.7$ \\
    3 & $\mathrm{MV_{R}}$ & $72.9$ & $92.1$ & $83.8$ & $92.7$\\
    \underline{4} & $\mathrm{DIS_{R}}$ & $74.1$ & $92.7$ & $84.7$ & $93.5$ \\
    5 & $\mathrm{RAFT}$ & $74.3$ & $93.2$ & $85.0$ & $93.8$ \\
    \bottomrule
\end{tabular}%

\end{table}
\PAR{Flow.} The effect of different optical flow methods used for warping is examined in \Cref{tab:ablations:flow}.
The tested approaches include motion vectors (MV) extracted from MPEG-4 video encoding, DIS~\cite{kroeger1026dis} flow, and SEA-RAFT~\cite{wang2024sea} flow.
Both MV and DIS flow can be used directly or refined via the "Flow Refine" convolutional layers described in \cref{ssec:method:design} and illustrated in \cref{fig:method:overview}, leading to refined versions $\mathrm{MV_R}$ and $\mathrm{DIS_R}$.
The results exhibit a diminishing return effect when increasing the quality of the optical flow $\mathbf{O}^t_{t-1}$, indicating that beyond a certain threshold, further improvements in flow estimation yield smaller gains.
Comparing row 1 to row 3 and row 2 to row 4, we observe that flow refinement, despite its relatively low capacity, improves both accuracy and consistency.
This suggests that the refinement step effectively denoises the warping flow, leading to better propagation.

\begin{table}[t]
\centering
\caption{\textbf{Propagation.} $\mathrm{Prop}$ refers to usage of propagation via flow-based warping, while $\mathrm{Init}$ to the Base Model initialization. $\mathrm{Enc_{Fast}}$ indicates which encoder is used for fast-frames, with fusion and refinement when applicable: ``Base (no prior)'' means the Base Encoder is used but no prior information is passed to.}
\vspace{-10pt}
\label{tab:ablations:propagation}
\resizebox{\linewidth}{!}{%
\begin{tabular}{cccc|cccc}
    \toprule
     &  Prop & Init & Enc\textsubscript{Fast} & $\mathrm{\delta_1}\uparrow$ & $\mathrm{\tau_5}\uparrow$ & $\mathrm{\delta^{SSI}_1}\uparrow$ & $\mathrm{\tau^{SSI}_5}\uparrow$ \\
    \midrule
    1 & \xmark & \xmark & Ours & $54.6$ & $74.5$ & $70.6$ & $82.2$ \\
    2 & \cmark & \xmark & Ours & $62.7$ & $90.7$ & $75.8$ & $92.4$ \\
    3 & \cmark & $-$ & Base (no prior) &  $78.1$ & $77.5$ & $86.3$ & $79.1$\\
    \underline{4} & \cmark & \cmark & Ours & $74.1$ & $92.7$ & $84.7$ & $93.5$ \\
    
    \bottomrule
    \end{tabular}%
}
\end{table}
\PAR{Propagation.} \Cref{tab:ablations:propagation} evaluates the role of initialization and propagation strategies.
Row 1 represents a standard image-based MDE model, where the PFW depth $\mathbf{D}^w_{t+1}$ and features  $\mathbf{F}^w_{t+1}$ are not utilized, thus we do not predict a residual but the full neck features $\mathbf{F}_t$ every frame.
Row 2 corresponds to \ourmodel without keyframe initialization from the Base Model,
Row 3 represents a model without any prior input modality but RGB (processed by the Base Encoder) and with flow-based propagation of previous neck features.
Comparing row 1 and row 2 highlights the importance of prior knowledge, framing the problem as a propagation rather than a prediction significantly improves both accuracy and consistency.
The comparison between row 2 and row 4 further emphasizes that a high-capacity initialization is highly beneficial for accuracy.
Additionally, the results in row 3 \vs row 4 show that while a high-capacity model enhances accuracy, it does not necessarily improve consistency.
Instead, the prior information from previous frames plays a crucial role in ensuring stable predictions.
This confirms that consistency is driven primarily by leveraging prior information rather than by increasing the capacity of the propagation mechanism alone.

\PAR{Keyframe Selection.} \cref{fig:method:kf_selection} illustrates the impact of different keyframe selection strategies.
We compare selecting keyframes at fixed intervals against our proposed heuristic described in \cref{ssec:method:propagation}.
Despite its simplicity and minimal tuning, our heuristic effectively maintains temporal consistency without sacrificing accuracy.
We note that increasing the distance between keyframes enhances consistency but negatively impacts accuracy, as the Propagation Module tends to produce overly smoothed results in the long run.
\section{Conclusion and Limitations}
\label{sec:conclusion}

We introduced a novel online video depth estimation approach leveraging temporal priors for enhanced consistency, efficiency, and accuracy: \ourmodel.
Our Propagation Module refines and propagates depth information across frames via optical flow and residual corrections, maintaining high temporal stability without computationally demanding recurrent architectures.
However, our approach relying on keyframe quality is fragile \wrt inaccuracies that may be propagated. 
Moreover, while the method's performance depends on optical flow input, results show robustness and flexibility.
Nonetheless, comprehensive zero-shot evaluations confirm that \ourmodel achieves superior temporal consistency and an optimal balance between accuracy, stability, and runtime efficiency, making it highly practical for real-world applications.

\vfill
\noindent{}\textbf{Acknowledgment.} This work is funded by Toyota Motor Europe via the research project TRACE-Z\"urich.

{
    \small
    \bibliographystyle{ieeenat_fullname}
    \bibliography{main}
}

\newpage

\maketitlesupplementary

\renewcommand\thesection{\Alph{section}}

This supplementary material offers further insights into our work.
We provide further quantitative results in \cref{sec:supp:quant}, specifically by providing an additional metric-depth evaluation in \Cref{tab:supp:comparison}, description of keyframe intervals defined by our keyframe selection mechanism introduced in Sec. 3.1, and test-time alternatives for the input flow in \Cref{tab:supp:flow}.
In \cref{sec:supp:faq}, we provide answers to possible questions that may arise.
Eventually, additional visualizations are provided in \cref{sec:supp:qual}.


\section{Additional Quantitative Results}
\label{sec:supp:quant}

\PAR{Metric evaluation.} \Cref{tab:supp:comparison} presents a comparison among methods providing metric depth predictions, including \ourmodel.
One notable strength of our model is its capability to deliver consistent and efficient video depth estimates directly in metric units, as its Base Model, \ie UnidepthV2. Our method substantially surpasses competing approaches in both consistency (+23.3\%) and efficiency (approximately $2\times$ faster), trailing only UniDepthV2 in terms of absolute accuracy.
However, not surpassing its Base Model highlights a limitation of the lightweight propagation module: while it efficiently maintains consistency, it does not inherently enhance depth accuracy.

\PAR{Test-time flow.} \Cref{tab:supp:flow} explores the impact of varying the quality of input flow estimations provided to \ourmodel at test time.
The results demonstrate that improving the initial flow accuracy does not necessarily translate to better overall performance (row 3), whereas reduced flow quality generally degrades depth predictions (row 1).
Notably, \ourmodel shows robustness against moderate noise increases in the input flow (row 1 \vs row 2).
Overall, these findings illustrate a diminishing return effect \wrt improving initial flow quality, consistent with observations previously discussed in Table 5.

\begin{figure}[t]
    \centering
    \footnotesize
    \includegraphics[width=1.0\linewidth]{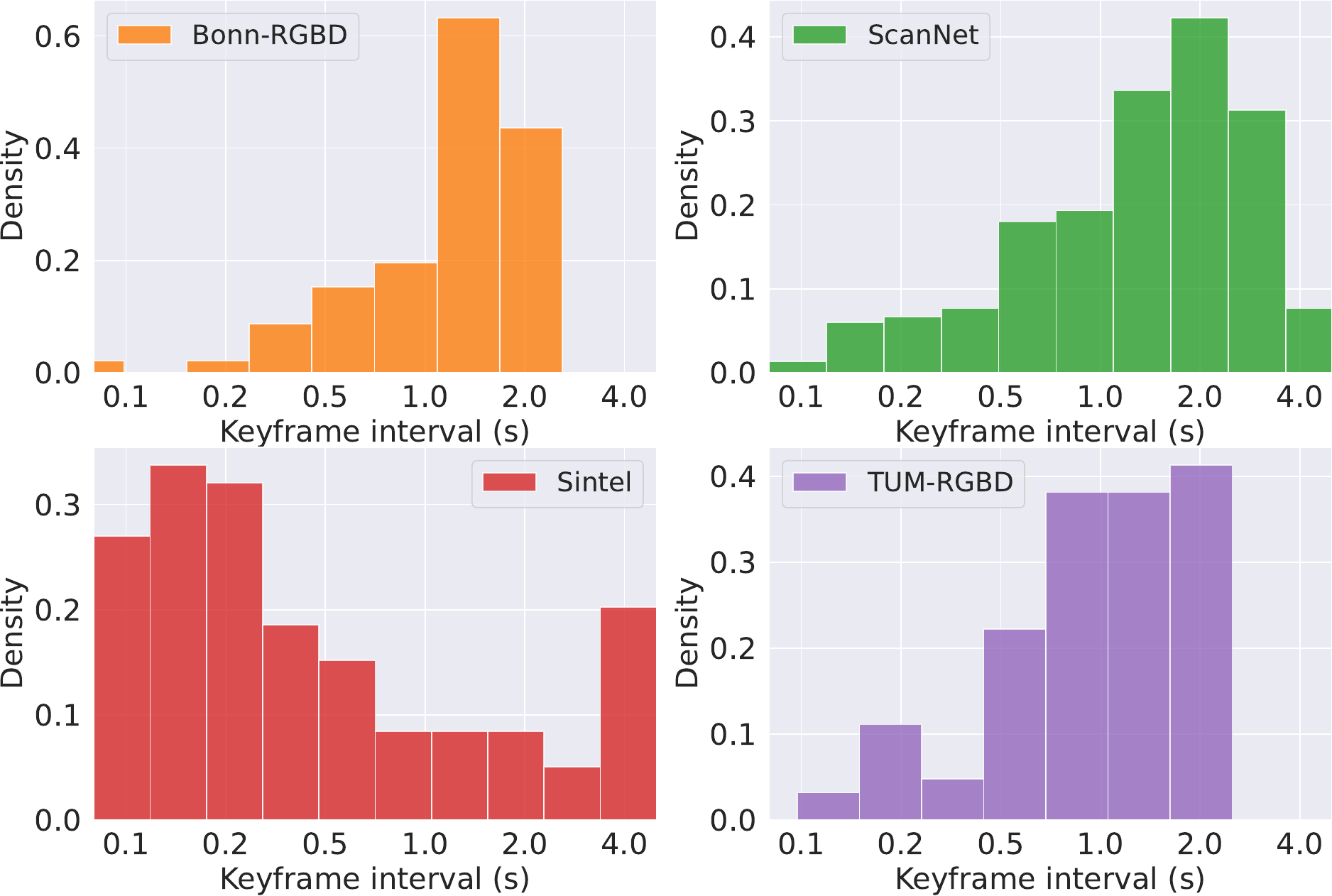}
    \vspace{-10pt}
    \caption{\textbf{Keyframe Distribution.} Our defined keyframe selection induces a distribution of intervals between keyframes induced for the four test sets, based on the nature and dynamics of each dataset.}
    \label{fig:method:kf_distrib}
\end{figure}

\PAR{Keyframe distribution.} The keyframe selection mechanism detailed in Sec. 3.1 offers enhanced flexibility by triggering new keyframes only when significant scene changes or strong motion occur.
Consequently, the interval between keyframes naturally adapts to the dynamics of each scene.
\Cref{fig:method:kf_distrib} illustrates the distribution of keyframe intervals across different datasets.
For instance, we observe shorter intervals in Sintel (mode around 0.18s), indicative of scenes with rapid motion, whereas ScanNet exhibits longer intervals (mode around 2.2s), reflecting lower dynamics.
We explore the impact of varying $\alpha,\beta,\gamma$ in \cref{tab:supp:kf_select}.
The results show how parameters impact the accuracy/consistency tradeoff but keep their average stable, confirming the algorithm's robustness.

\begin{table*}[t]
\centering
\caption{\textbf{Comparison on zero-shot evaluation for metric predictions.} All methods are evaluated in an online fashion. The ``Type'' column indicates the original task tackled, \textbf{M}: monocular, \textbf{S}: stereo, \textbf{V}: video. Profiling is run on 60-frame 0.5MP clip averaged per frame. $\mathrm{FPS_{GPU}}$ is measured on an RTX 3090 and $\mathrm{FPS_{CPU}}$ on an M1 chip, both with half precision.}
\vspace{-10pt}
\label{tab:supp:comparison}
\resizebox{\linewidth}{!}{%
\begin{tabular}{lccccccccccccccc}
\toprule
\multirow{2}{*}{\textbf{Method}} & \multirow{2}{*}{\textbf{Type}} & \multicolumn{2}{c}{TUM-RGBD} & \multicolumn{2}{c}{ScanNet} & \multicolumn{2}{c}{Sintel} & \multicolumn{2}{c}{Bonn-RGBD} & \multicolumn{2}{c}{\textbf{Aggregate}} & \multicolumn{2}{c}{\textbf{Efficiency}} \\
\cmidrule(lr){3-4} \cmidrule(lr){5-6} \cmidrule(lr){7-8} \cmidrule(lr){9-10} \cmidrule(lr){11-12} \cmidrule(lr){13-14}
 &  & $\mathrm{\delta_1}\uparrow$ & $\mathrm{\tau_{5}}\uparrow$ & $\mathrm{\delta_1}\uparrow$ & $\mathrm{\tau_{5}}\uparrow$ & $\mathrm{\delta_1}\uparrow$ & $\mathrm{\tau_{5}}\uparrow$ & $\mathrm{\delta_1}\uparrow$ & $\mathrm{\tau_{5}}\uparrow$ & $\mathrm{\delta_1}\uparrow$ & $\mathrm{\tau_{5}}\uparrow$ & $\mathrm{FPS_{GPU}}\uparrow$ & $\mathrm{FPS_{CPU}}\uparrow$ \\
\midrule
Metric3Dv2~\cite{hu2024metric3dv2} & M & $81.2$ & $\scnd{81.2}$ & $82.0$ & $\scnd{88.5}$ & $\best{35.6}$ & $41.0$ & $96.9$ & $\scnd{92.6}$ & $71.3$ & $76.2$ & $7.1$ & $0.7$ \\
DepthPro~\cite{bochkovskii2024depthpro} & M & $66.4$ & $67.5$ & $60.7$ & $59.6$ & $30.0$ & $26.3$ & $92.4$ & $77.2$ & $61.0$ & $58.0$ & $2.8$ & $0.2$ \\
UniDepthV2~\cite{piccinelli2025unidepthv2} & M & $\best{88.1}$ & $79.7$ & $\best{96.2}$ & $85.7$ & $\scnd{35.3}$ & $\scnd{52.2}$ & $\best{97.7}$ & $90.4$ & $\best{79.8}$ & $\scnd{77.4}$ & $\scnd{13.4}$ & $\scnd{1.0}$ \\
MASt3R~\cite{leroy2024master} & S & $35.7$ & $45.6$ & $54.4$ & $60.5$ & $16.1$ & $44.8$ & $63.7$ & $69.8$ & $43.1$ & $54.4$ & $2.7$ & $0.7$ \\
\midrule
\ourmodel & V & $\scnd{82.9}$ & $\best{94.9}$ & $\scnd{91.9}$ & $\best{91.4}$ & $33.1$ & $\best{80.3}$ & $\scnd{97.3}$ & $\best{96.3}$ & $\scnd{76.3}$ & $\best{90.7}$ & $\best{26.2}$ & $\best{2.7}$ \\
\bottomrule
\end{tabular}%
}
\end{table*}

\begin{table}[ht]
\centering
\caption{\textbf{Test-time flow.} $\mathbf{O}^{t}_{t-1}$ is the flow used to perform warping during test-time. $\mathrm{MV}$ refers to using MPEG-4 motion vectors, $\mathrm{DIS}$ utilizes~\cite{kroeger1026dis} and $\mathrm{RAFT}$~\cite{wang2024sea}. The input flow is always refined via $\mathrm{FlowRefine}$. The model is restored from the fully trained checkpoint.}
\vspace{-10pt}
\label{tab:supp:flow}
\begin{tabular}{cc|cccc}
    \toprule
     & $\mathbf{O}^{t}_{t-1}$ & $\mathrm{\delta_1}\uparrow$ & $\mathrm{\tau_5}\uparrow$ & $\mathrm{\delta^{SSI}_1}\uparrow$ & $\mathrm{\tau^{SSI}_5}\uparrow$  \\
    \midrule
    1 & $\mathrm{MV}$ & $73.5$ & $88.2$ & $90.0$ & $85.8$\\
    \underline{2} & $\mathrm{DIS}$ & $76.3$ & $90.7$ & $91.3$ & $87.3$ \\
    3 & $\mathrm{RAFT}$ & $76.3$ & $90.7$ & $91.4$ & $87.2$ \\
    \bottomrule
\end{tabular}%

\end{table}

\begin{figure}[t]
    \renewcommand{\arraystretch}{1}
    \centering
    \small
    \begin{tabular}{cc}
        \includegraphics[width=0.45\linewidth]{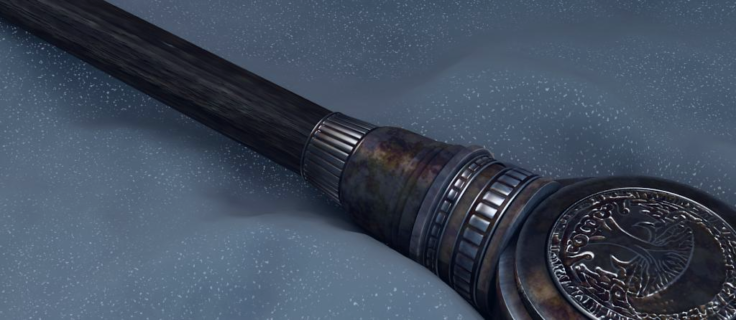}
        & \includegraphics[width=0.45\linewidth]{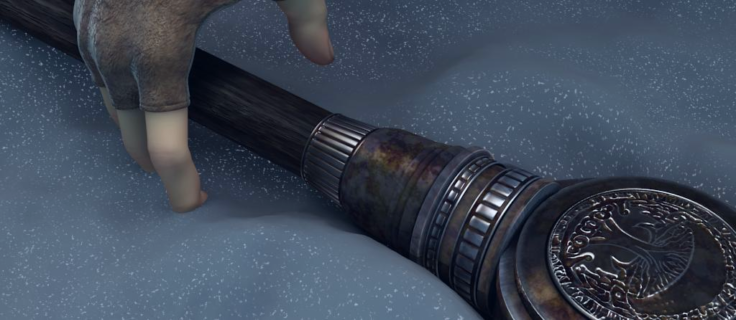}\\
        \includegraphics[width=0.45\linewidth]{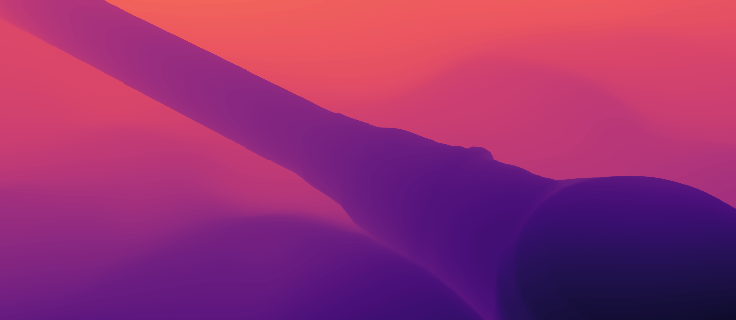}
        & \includegraphics[width=0.45\linewidth]{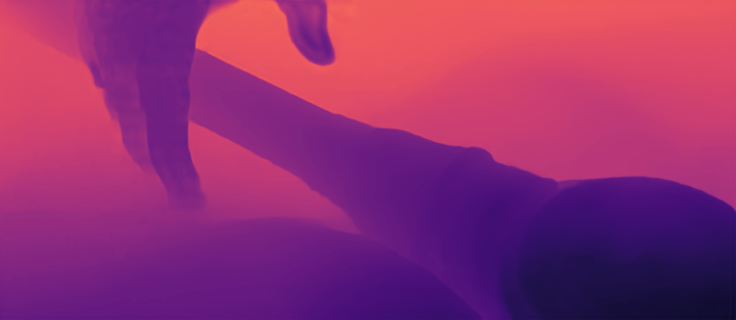}\\
    \end{tabular}
    \vspace{-10pt}
    \caption{\textbf{New Object in Sequence.} The left two images show the keyframe and the predicted depth, and the right ones display the RGB and the corresponding depth prediction after ten consecutive propagations. The hand on the right is a newly appearing object introduced during propagation.}
    \label{fig:supp:occ}
\end{figure}

\begin{table*}[ht]
\centering
\caption{\textbf{Online vs. Offline video depth.} The first column indicates the evaluation regime: \emph{Offline}: full-sequence access at any timestep; \emph{Online}: no future frames are available for the current timestep}
\label{tab:supp:offline}
\resizebox{\linewidth}{!}{%
\begin{tabular}{llcccccccccc}
\toprule
\multirow{2}{*}{} & \multirow{2}{*}{\textbf{Method}} & \multicolumn{2}{c}{TUM-RGBD} & \multicolumn{2}{c}{ScanNet} & \multicolumn{2}{c}{Sintel} & \multicolumn{2}{c}{Bonn-RGBD} & \multicolumn{2}{c}{\textbf{Aggregate}} \\
\cmidrule(lr){3-4} \cmidrule(lr){5-6} \cmidrule(lr){7-8} \cmidrule(lr){9-10} \cmidrule(lr){11-12}
 &  & $\mathrm{\delta^{SSI}_1}\uparrow$ & $\mathrm{\tau^{SSI}_{5}}\uparrow$ & $\mathrm{\delta^{SSI}_1}\uparrow$ & $\mathrm{\tau^{SSI}_{5}}\uparrow$ & $\mathrm{\delta^{SSI}_1}\uparrow$ & $\mathrm{\tau^{SSI}_{5}}\uparrow$ & $\mathrm{\delta^{SSI}_1}\uparrow$ & $\mathrm{\tau^{SSI}_{5}}\uparrow$ & $\mathrm{\delta^{SSI}_1}\uparrow$ & $\mathrm{\tau^{SSI}_{5}}\uparrow$ \\
\midrule
\multirow{4}{*}{\rotatebox{90}{Offline}} 
 & ChronoDepth~\cite{shao2024chrono} & $68.5$ & $81.4$ & $79.2$ & $81.1$ & $35.3$ & $43.7$ & $64.2$ & $74.1$ & $61.8$ & $70.1$ \\
 & DepthCrafter~\cite{hu2024crafter} & $88.8$ & $87.3$ & $96.1$ & $89.6$ & $\best{79.8}$ & $70.5$ & $\best{99.0}$ & $\scnd{96.9}$ & $90.0$ & $84.8$ \\
 & VideoDA~\cite{chen2025vda} & $\best{95.1}$ & $83.2$ & $\scnd{98.2}$ & $\scnd{90.3}$ & $76.9$ & $\scnd{71.4}$ & $\best{99.0}$ & $96.6$ & $\best{92.2}$ & $84.4$ \\
 & VGG-T~\cite{wang2025vggt} & $91.6$ & $\best{91.1}$ & $\best{99.4}$ & $\best{91.6}$ & $75.8$ & $\best{76.2}$ & $99.0$ & $\best{96.9}$ & $91.6$ & $\best{88.4}$\\
\midrule
\multirow{5}{*}{\rotatebox{90}{Online}} 
 & ChronoDepth~\cite{shao2024chrono} & $58.5$ & $76.6$ & $74.1$ & $72.4$ & $26.5$ & $34.2$ & $61.8$ & $68.8$ & $55.2$ & $63.0$ \\
 & DepthCrafter~\cite{hu2024crafter} & $85.6$ & $80.0$ & $94.5$ & $83.2$ & $\scnd{77.3}$ & $64.9$ & $\scnd{98.9}$ & $96.3$ & $88.0$ & $79.2$ \\
 & VideoDA~\cite{chen2025vda} & $\scnd{94.9}$ & $80.1$ & $98.1$ & $87.5$ & $76.8$ & $64.2$ & $\best{99.0}$ & $96.0$ & $\best{92.2}$ & $80.7$ \\
 & VGG-T~\cite{wang2025vggt} & $90.2$ & $84.4$ & $\scnd{99.3}$ & $86.5$ & $74.1$ & $68.9$ & $99.0$ & $96.6$ & $90.9$ & $83.0$ \\
 & \ourmodel & $94.8$ & $\scnd{89.1}$ & $96.2$ & $89.1$ & $76.1$ & $\best{75.4}$ & $98.4$ & $96.3$ & $\scnd{91.4}$ & $\best{87.5}$ \\
\bottomrule
\end{tabular}%
}
\end{table*}

\begin{table}[t]
\centering
\setlength{\tabcolsep}{3pt}
\footnotesize
\renewcommand{\arraystretch}{0.9}
\caption{\textbf{Keyframe Selection.} Metrics: $\mathrm{\delta^{SSI}_1}\uparrow / \mathrm{\tau^{SSI}_{5}}\uparrow$}
\vspace{-10pt}
\label{tab:supp:kf_select}
\resizebox{\linewidth}{!}{%
\begin{tabular}{ccccccc}
\toprule
$\alpha$ & $\beta$ & $\gamma$ & TUM & ScanNet & Sintel & Bonn\\
\midrule
$0.00$ & $0.10$ & $0.20$ & $96.5$ / $86.3$ & $98.4$ / $81.8$ & $78.8$ / $67.6$ & $99.0$ / $94.9$ \\
$0.00$ & $0.15$ & $0.20$ & $96.4$ / $86.6$ & $98.2$ / $84.0$ & $78.3$ / $70.2$ & $98.9$ / $95.3$ \\
$0.10$ & $0.10$ & $0.20$ & $94.7$ / $89.1$ & $96.1$ / $89.3$ & $75.2$ / $75.6$ & $98.4$ / $96.2$ \\
$0.10$ & $0.10$ & $0.25$ & $92.8$ / $90.0$ & $95.4$ / $89.8$ & $75.1$ / $75.5$ & $97.6$ / $97.4$ \\
$\underline{0.10}$ & $\underline{0.15}$ & $\underline{0.20}$ & $94.8$ / $89.2$ & $95.6$ / $89.8$ & $75.9$ / $76.5$ & $98.4$ / $96.2$ \\
$0.10$ & $0.15$ & $0.25$ & $92.7$ / $90.1$ & $94.9$ / $90.4$ & $75.8$ / $76.5$ & $97.6$ / $97.4$ \\
\bottomrule
\end{tabular}
} 
\end{table}

\section{Q\&A}
\label{sec:supp:faq}

In this section, we address potential questions and points of curiosity readers may have after reviewing the paper. The section is organized in a question-and-answer format.

\begin{itemize}

\item \textbf{Does the proposed metric $\tau_5$, which relies on warping, introduce evaluation noise?} \\ Yes. Warping inevitably introduces some errors due to resampling, even when applied using ground truth (GT) depth and flow. For example, applying $\tau_5$ with GT typically yields scores above $95\%$ but not a perfect $100\%$. However, if egomotion compensation is omitted, $\tau_5$ on GT drops below $80\%$. This significant gap demonstrates the superior informativeness of our metric compared to previous metrics.

\item \textbf{Can the model handle newly appearing objects?} \\ Yes, \ourmodel can effectively manage significant scene changes, including the appearance or disappearance (occlusions/disocclusions) of objects as in \cref{fig:supp:occ}. Nonetheless, when a large motion dominates the scene, the accuracy of propagation tends to degrade.

\item \textbf{Isn't 2.7 FPS on a mobile device insufficient for real-time applications?} \\
True. However, the reported 2.7 FPS is for relatively high-resolution (0.5 MP) images. For true real-time performance, lowering the resolution to a more mobile-friendly size, such as $640 \times 360$, increases the speed significantly—to over 60 FPS on a GPU and around 10 FPS on mobile devices. This performance is sufficient for moderately dynamic scenarios, such as the ScanNet evaluation, conducted at 10 FPS.

\item \textbf{Your efficiency improvements target dynamic, real-time scenarios, yet performance deteriorates in highly dynamic scenes, or it requires the slower Base Model. Isn't this contradictory?} \\
As discussed in Sec. 3.2, \ourmodel assumes relatively small changes between frames, allowing it to correct regions rather than predict from scratch. In scenarios with significant scene dynamics, this assumption no longer holds, leading to reduced accuracy. One practical solution is to increase input FPS to reduce apparent motion and flow magnitude. However, this remains an inherent limitation common to all propagation methods. For instance, even standard video encoders typically reduce bitrate or resolution to handle high-motion content.

\item \textbf{Propagation tasks are not novel. Why do you not reference methods from Video Object Segmentation/Detection (VOS/VOD)?} \\
Initialization-based propagation methods are indeed widely explored in Video Object Segmentation/Detection (VOS/VOD), which focus on tracking segmentation masks or bounding boxes across frames~\cite{perazzi2016benchmark}. However, video depth propagation fundamentally differs as it involves dense pixel-wise regression rather than discrete classification or segmentation. Depth propagation requires accurate predictions for all pixels, unlike VOS/VOD methods, which track only specific object subsets. Moreover, most VOS methods are inherently offline, rendering them impractical for our real-time propagation task.

\item \textbf{How do you calculate FPS given \ourmodel's adaptive keyframe interval?} \\
We compute the average keyframe interval for a given scene, typically spanning approximately 30 frames. The FPS is then calculated over a 60-frame clip, assuming a keyframe is triggered every 30 frames on average.

\item \textbf{Why does \ourmodel use only UniDepthV2? Can it work with other base models?} \\ Yes, \ourmodel can be paired with any base model capable of providing generalizable metric 3D output, including depth (or distance) and camera parameters. Currently, the most popular methods to fit this criterion are UniDepth (V1 and V2) and DepthPro. We chose UniDepthV2 due to its accessibility and ease of training. Notably, the relative efficiency gain would be even larger with DepthPro, as its decoder is similar to UniDepthV2’s, while its encoder is significantly heavier. Moreover, being paired with 3D estimation allows \ourmodel to be able to use the camera predictions from the Base Model's keyframe and actually output a 3D point cloud.

\item \textbf{Does refinement improve the flow $\mathbf{O}^{t}_{t-1}$?} \\ Yes, refinement enhances flow accuracy. For example, DIS flow has an End-Point Error (EPE) of 8.5px compared to SEA-RAFT, which refinement reduces to 5.2px. Similarly, motion vectors (MVs) start with an EPE of 12.8px, which refinement lowers to 7.3px.

\item \textbf{Why is generally $\tau^{\mathrm{SSI}}_5$ worse than $\tau_5$ ?} \\ While $\tau^{\mathrm{SSI}}_5$ removes scale and shift relative to the ground truth by applying an affine transformation, it can introduce higher inconsistency. Local jittering may lead to large and inconsistent residuals when estimating the affine transformation via least squares, resulting in temporally incoherent per-frame global statistics that can actually exacerbate the inconsistency.

\item \textbf{Should offline self-attention be the upper-bound?} \\ Self‑attention helps mostly when future frames are visible. When run causally, DepthCrafter and VideoDA drop $7–10\%$ in $\tau^{SSI}_5$ (\cref{tab:supp:offline}, ``online''), and VeloDepth overtakes both on Sintel and TUM.  The ``upper bound'' is thus offline access, not self-attention itself. Also, DepthCrafter leans on 3D convs rather than heavy temporal attention.
Anyway, we do not claim to beat their offline scores, although we partially match, but we exceed their online consistency at real‑time speed as shown in \Cref{tab:supp:offline}.

\item \textbf{Why some reported results differ, both postively and negatively, from the results reported in original papers?} \\  Baselines output either disparity or depth: we compute scale–shift, one per sequence, in that same domain to avoid train‑test mismatch, and we mask sky pixels and out‑of‑range backgrounds.
This consistent post‑processing might raise accuracies beyond the originals while keeping the comparison fair.

\item \textbf{How is efficiency computed?} \\ We utilized DeepSpeed's tools to calculate FLOPs and total parameters in \Cref{tab:results:comparison}.
We omit GPU memory numbers because they depend on kernel choices and optimization (\eg FlashAttention); parameter count is therefore the more hardware‑agnostic measure.
\ourmodel uses 409M parameters in total, only 55M for propagation, and needs 5 to 38$\times$ fewer FLOPs than recent video depth baselines.

\end{itemize}

\section{Additional Qualitative Results}
\label{sec:supp:qual}
\begin{figure*}[ht]
    \renewcommand{\arraystretch}{1}
    \centering
    \small
    \begin{tabular}{ccc|ccc|c}
        \multirow{1}{*}[2.9em]{\rotatebox[origin=c]{90}{Bonn}}
        & \includegraphics[width=0.15\linewidth]{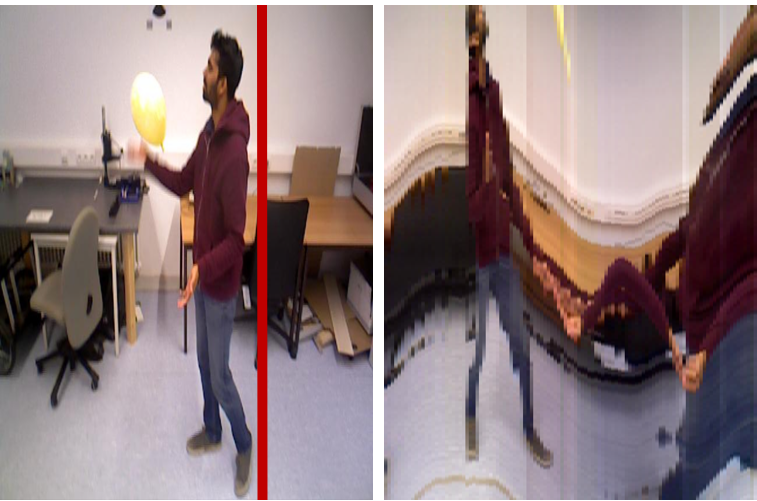}
        & \includegraphics[width=0.15\linewidth]{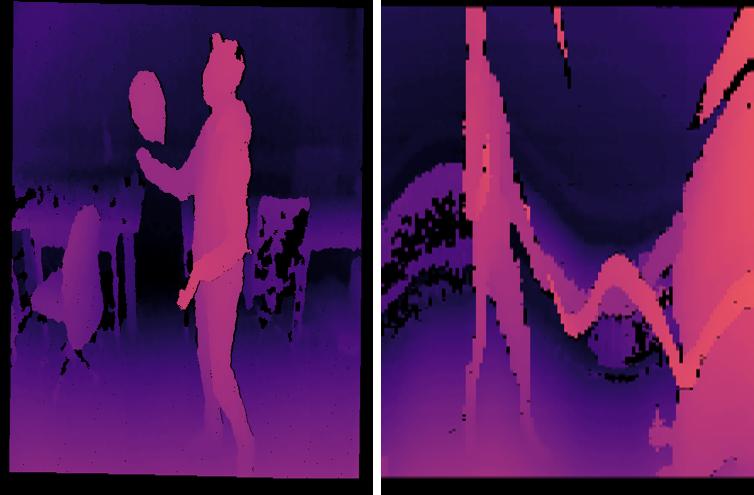}
        & \includegraphics[width=0.15\linewidth]{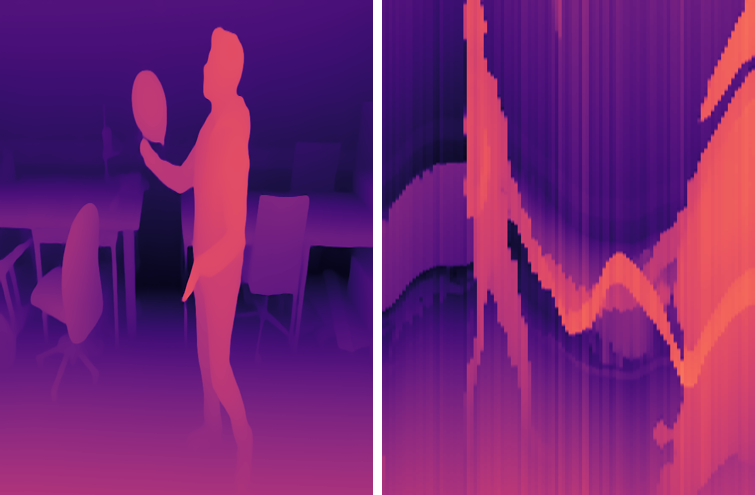}
        & \includegraphics[width=0.15\linewidth]{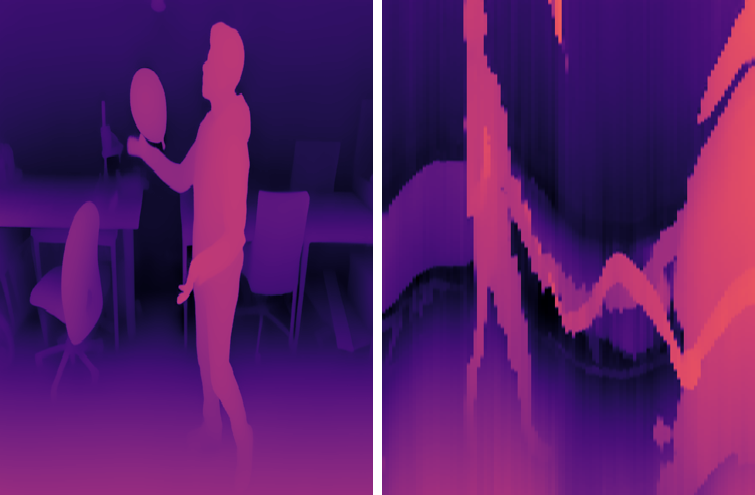}
        & \includegraphics[width=0.15\linewidth]{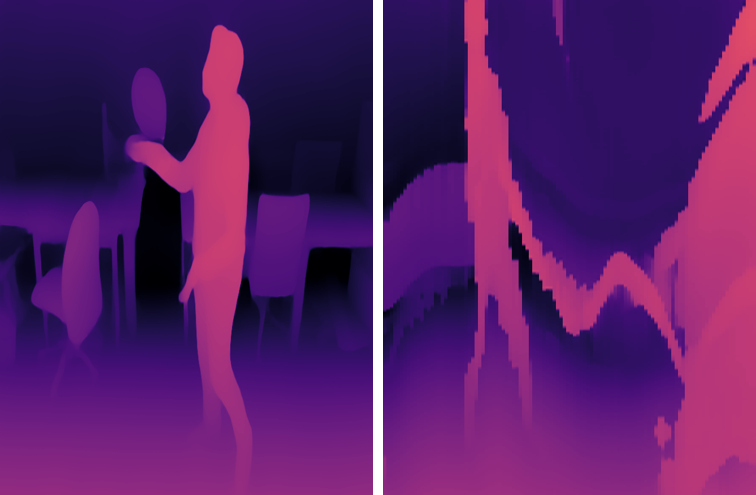}
        & \includegraphics[width=0.025\linewidth]{figures/qualitative/colorbar_8.png}\\

        \multirow{1}{*}[3.0em]{\rotatebox[origin=c]{90}{TUM}}
        & \includegraphics[width=0.15\linewidth]{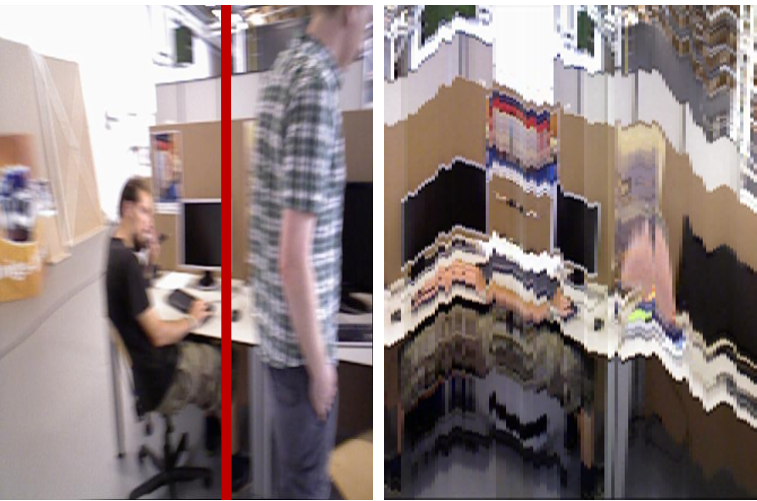}
        & \includegraphics[width=0.15\linewidth]{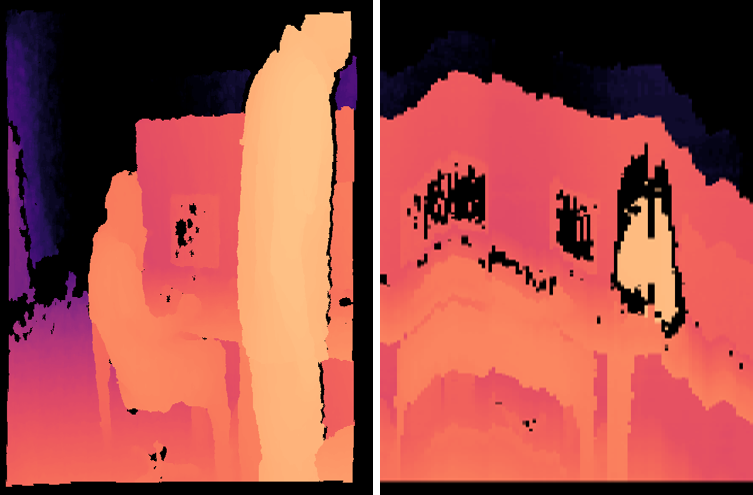}
        & \includegraphics[width=0.15\linewidth]{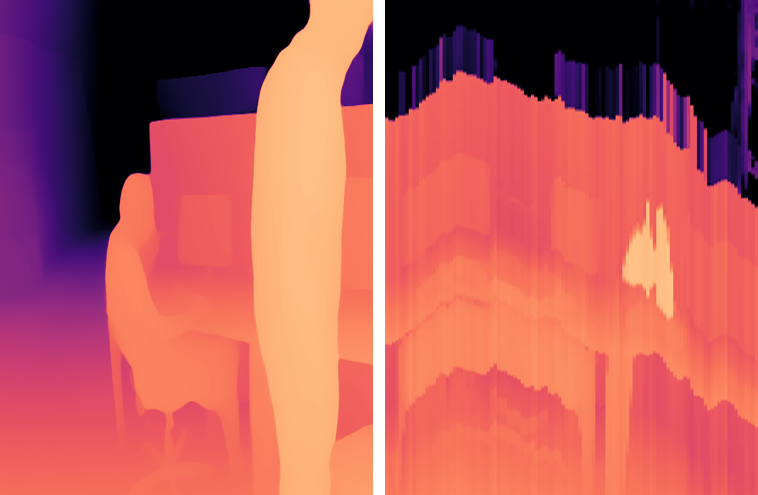}
        & \includegraphics[width=0.15\linewidth]{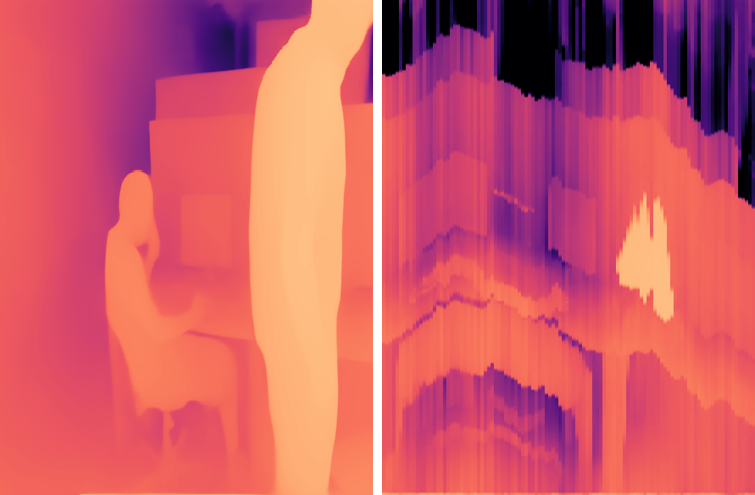}
        & \includegraphics[width=0.15\linewidth]{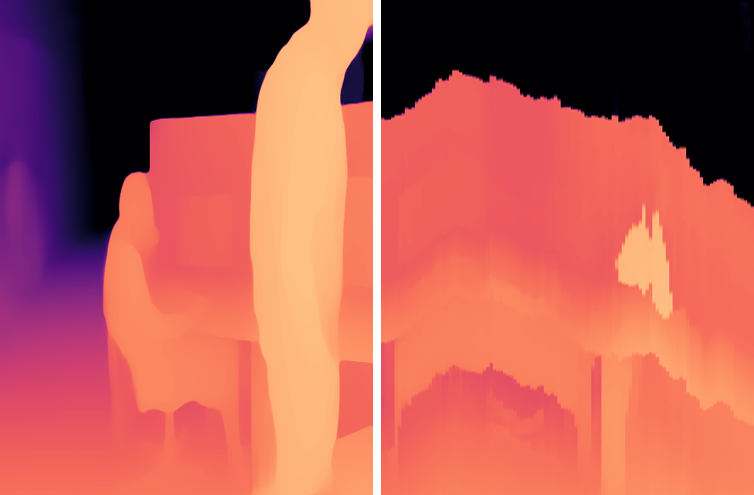}
        & \includegraphics[width=0.025\linewidth]{figures/qualitative/colorbar_8.png}\\

        \multirow{1}{*}[3.0em]{\rotatebox[origin=c]{90}{Sintel}}
        & \includegraphics[width=0.15\linewidth]{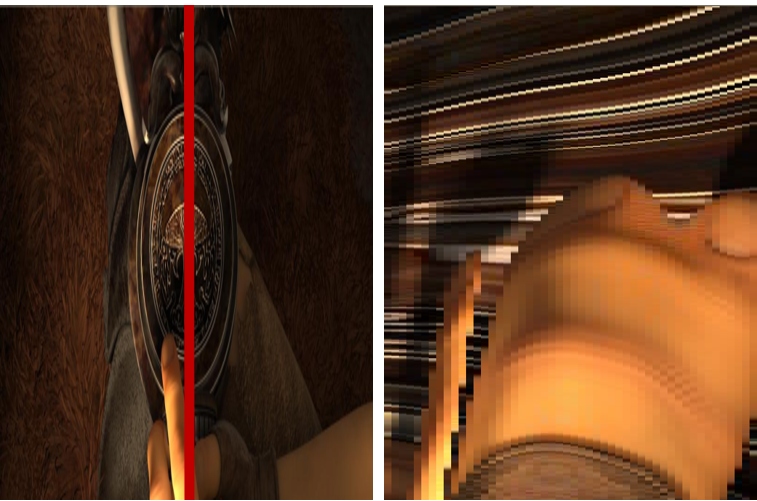}
        & \includegraphics[width=0.15\linewidth]{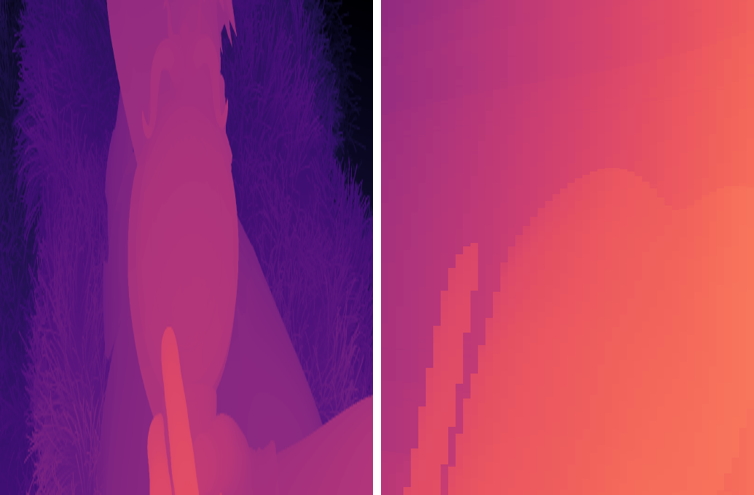}
        & \includegraphics[width=0.15\linewidth]{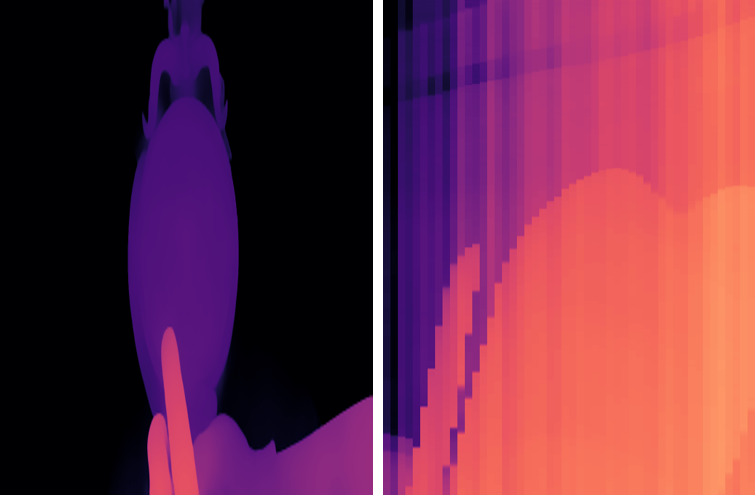}
        & \includegraphics[width=0.15\linewidth]{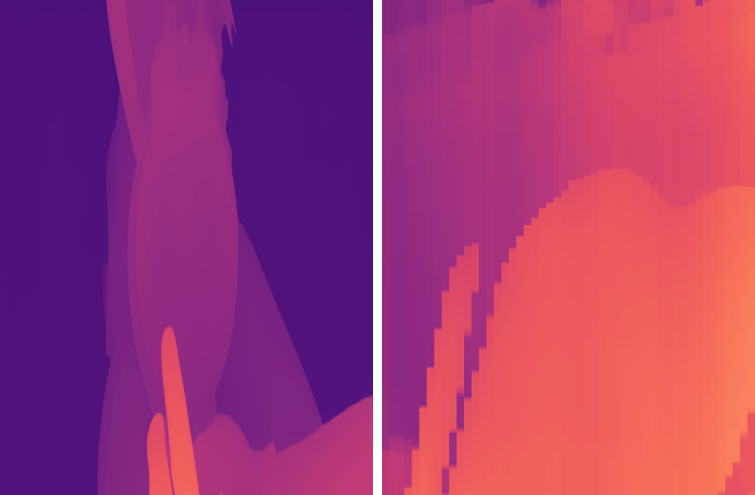}
        & \includegraphics[width=0.15\linewidth]{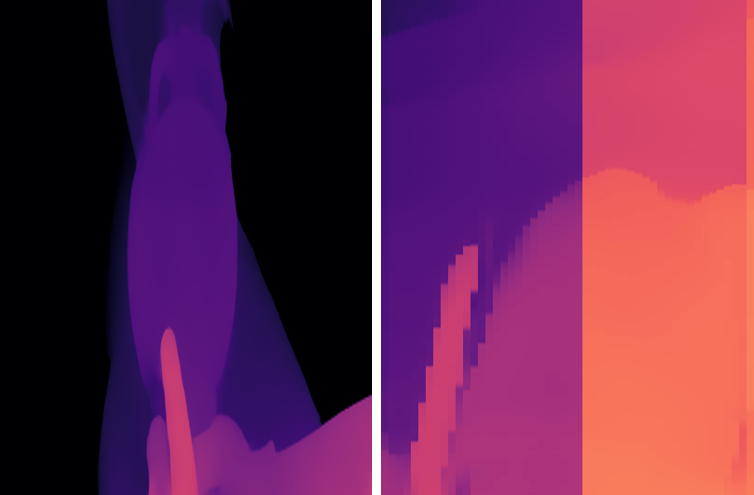}
        & \includegraphics[width=0.025\linewidth]{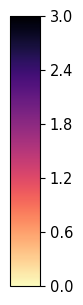}\\

        \multirow{1}{*}[3.5em]{\rotatebox[origin=c]{90}{ScanNet}}
        & \includegraphics[width=0.15\linewidth]{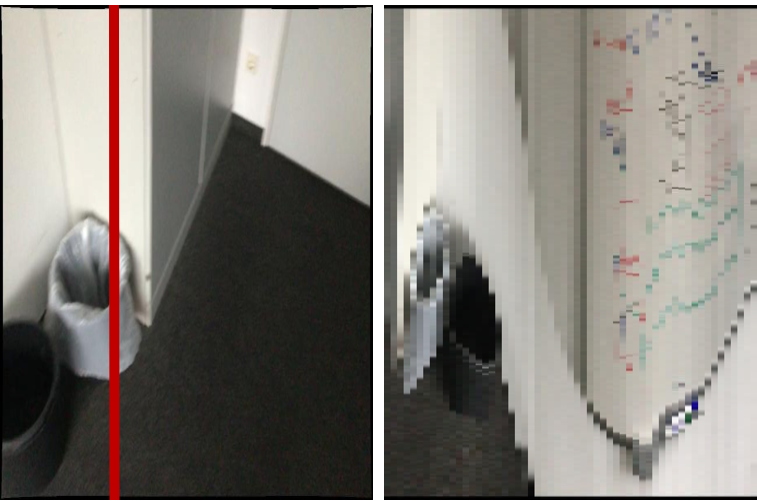}
        & \includegraphics[width=0.15\linewidth]{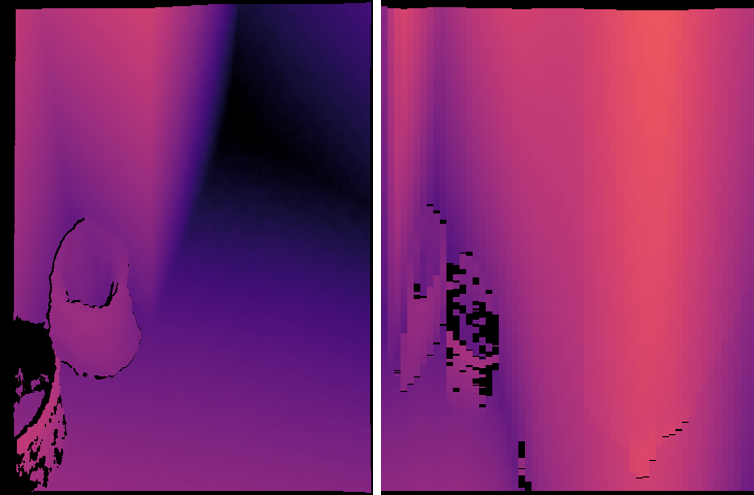}
        & \includegraphics[width=0.15\linewidth]{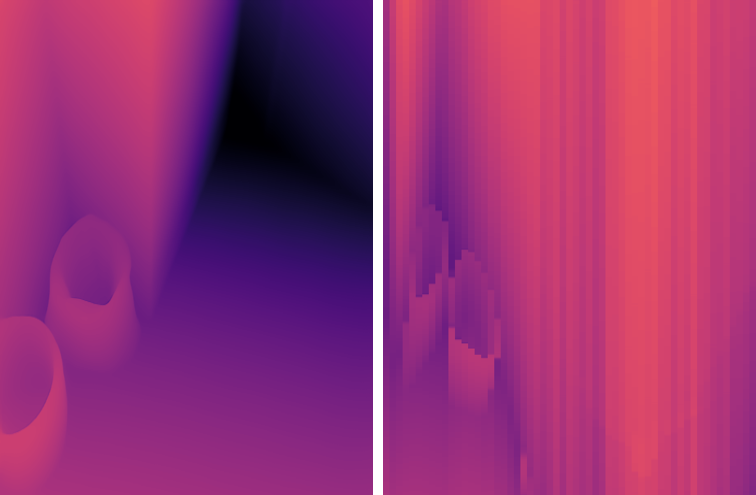}
        & \includegraphics[width=0.15\linewidth]{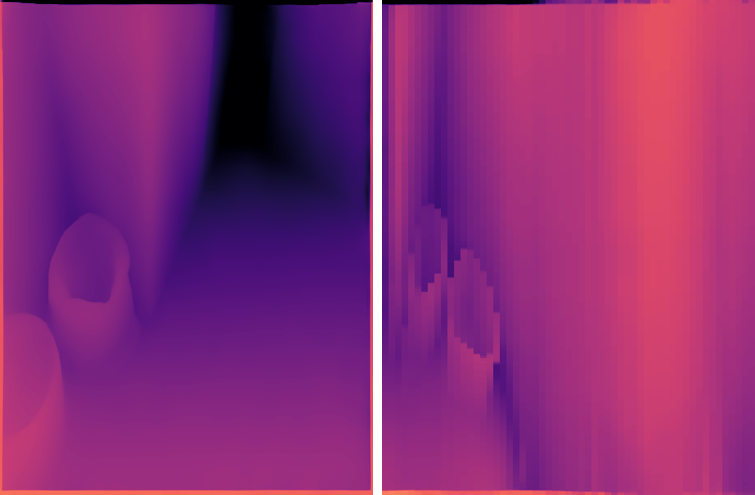}
        & \includegraphics[width=0.15\linewidth]{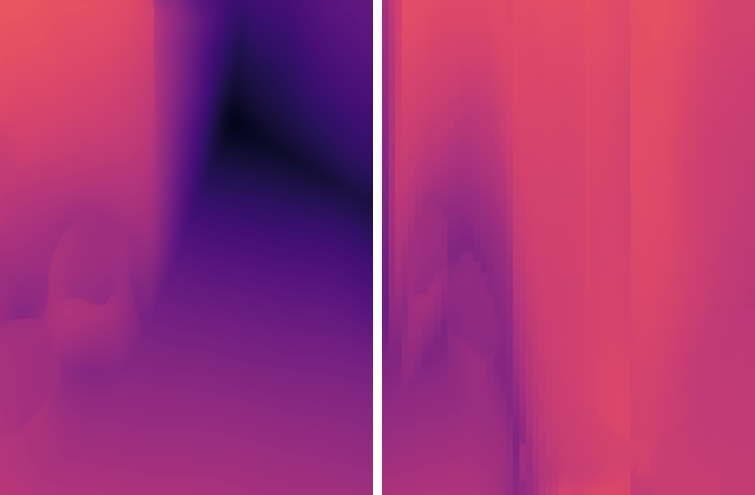}
        & \includegraphics[width=0.025\linewidth]{figures/qualitative/colorbar_5.png}\\
    
        & RGB & GT & UniK3D~\cite{piccinelli2025unik3d} & DepthCrafter\textsuperscript{\dag}~\cite{hu2024crafter} & \ourmodel & Depth \\
    \end{tabular}
    \vspace{-10pt}
    \caption{\textbf{Zero-shot qualitative results.} Each row corresponds to one test video sample from one domain. Each block shows the 6th frame and the video slices corresponding to the red line x-location in the first column. UniDepthV2 and \ourmodel outputs are inherently metric. No post-processing is applied. The last column represents the depth values \wrt ``magma'' colormap. (\dag): affine transformed to match GT. Best viewed on a screen and zoomed in.}
    \label{fig:supp:main_vis}
\end{figure*}

We provide here more qualitative comparisons, particularly additional dynamic scenes not present in the main paper, are reported in \cref{fig:supp:main_vis}.
The visualization clearly shows improved consistency \wrt the Base Model or video depth estimation, but the ``keyframe jump'' is still present in, \eg Sintel's close-cut visualization, as detailed in Sec. 4.2; nonetheless, simple global statistic smoothing, \ie Exponential Moving Average, over time may reduce the impact.
In addition, we test our model on test datasets and in-the-wild scenarios in the video supplements and visualize the corresponding video point clouds without postprocessing.
The visualizations presented here, both from the validation sets and the in-the-wild ones are casually selected and not cherry-picked.


\end{document}